\documentclass{article}

\usepackage[preprint]{neurips_2019}
\usepackage{rotating}
\usepackage{placeins}
\usepackage[export]{adjustbox}
\usepackage{lipsum} 

\usepackage{graphicx}
\usepackage{xparse}

\usepackage{color}
\usepackage{titlesec}
\usepackage{wrapfig}

\usepackage[font=small,labelfont=bf]{caption}

\usepackage{graphicx}
\usepackage{pgfplots}
\pgfplotsset{compat=1.11}

\usepackage{float}

\usepackage{morefloats}
\usepackage{mathtools}

\usepackage{placeins}
\usepackage{paralist}

\usepackage{xspace}

\usepackage{multirow}

\usepackage{ltablex}

\usepackage{booktabs}

\newcolumntype{Y}{>{\centering\arraybackslash}X}

\usepackage{caption} 
\usepackage[inline]{enumitem}

\usepackage{algorithm}
\usepackage{algorithmicx}
\usepackage{algpseudocode}
\usepackage{amsmath,amsfonts,amsthm,bm} 
\usepackage{mathtools}
\usepackage{mathrsfs}

\usepackage[title]{appendix}

\usepackage[utf8]{inputenc} 
\usepackage[T1]{fontenc}    
\usepackage{hyperref}       
\usepackage{url}            
\usepackage{booktabs}       
\usepackage{amsfonts}       
\usepackage{nicefrac}       
\usepackage{microtype}      
\usepackage{tikz}
\usepackage{gensymb}
\usepackage{caption}
\usepackage{subcaption}

\usepackage{defs}
\usepackage{xargs}  
\usepackage{soul}
\usepackage{lipsum} 
\usepackage{xcolor}
\usepackage{enumitem}
\setlist[itemize]{noitemsep,topsep=-6pt}
\setlist[enumerate]{noitemsep,topsep=-6pt}

\usepackage{nndefs}

\usetikzlibrary{shapes,arrows}

\DeclarePairedDelimiterX{\infdivx}[2]{(}{)}{%
  #1\;\delimsize|\delimsize|\;#2%
}

\DeclareMathOperator*{\E}{\mathbb{E}}

\author{
  Gautam Singh\thanks{Equal contribution
  } \\
  Rutgers University\\
  \texttt{singh.gautam@rutgers.edu}
  \And
  Jaesik Yoon\footnotemark[1] \\
  SAP\\
  \texttt{jaesik.yoon01@sap.com}
  \And
  Youngsung Son \\
  ETRI\\
  \texttt{ysson@etri.re.kr}
  \And
  Sungjin Ahn \\
  Rutgers University\\
  \texttt{sungjin.ahn@rutgers.edu}
}

\usepackage{xargs}  
\usepackage{xcolor}
\usepackage[colorinlistoftodos,textsize=tiny,textwidth=3.4cm]{todonotes}
\newcommandx{\ahn}[2][1=]{\todo[linecolor=red,backgroundcolor=red!5,bordercolor=red,#1]{#2}}
\newcommandx{\gautam}[2][1=]{\todo[linecolor=blue,backgroundcolor=blue!5,bordercolor=blue,#1]{#2}}
\newcommandx{\jaesik}[2][1=]{\todo[linecolor=green,backgroundcolor=green!5,bordercolor=green,#1]{#2}}

\begin{document}
\tikzstyle{block} = [draw, fill=blue!20, rectangle, 
    minimum height=3em, minimum width=6em]
\tikzstyle{sum} = [draw, fill=blue!20, circle, node distance=1cm]
\tikzstyle{input} = [coordinate]
\tikzstyle{output} = [coordinate]
\tikzstyle{pinstyle} = [pin edge={to-,thin,black}]

\title{Sequential Neural Processes}

\maketitle

\begin{abstract}
Neural Processes combine the strengths of neural networks and Gaussian processes to achieve both flexible learning and fast prediction in stochastic processes. However, a large class of problems comprises underlying temporal dependency structures in a sequence of stochastic processes that Neural Processes (NP) do not explicitly consider.~In this paper, we propose Sequential Neural Processes (SNP) which incorporates a temporal state-transition model of stochastic processes and thus extends its modeling capabilities to dynamic stochastic processes. In applying SNP to dynamic 3D scene modeling, we introduce the Temporal Generative Query Networks. To our knowledge, this is the first 4D model that can deal with the temporal dynamics of 3D scenes. In experiments, we evaluate the proposed methods in dynamic (non-stationary) regression and 4D scene inference and rendering.
\end{abstract}

\section{Introduction}

Neural networks consume all training data and computation through a costly training phase to engrave a single function into its weights.~While this makes us entertain fast prediction on the learned function, under this rigid regime changing the target function means costly retraining of the network.~This lack of flexibility thus plays as a major obstacle in tasks such as meta-learning and continual learning where the function needs to be changed over time or on-demand.~Gaussian processes (GP) do not suffer from this problem.~Conditioning on observations, it directly performs inference on the target stochastic process.~Consequently, Gaussian processes show the opposite properties to neural networks:~it is flexible in making predictions because of its non-parametric nature,
but this flexibility comes at a cost of having slow prediction.~GPs can also capture the uncertainty on the estimated function. 

Neural Processes (NP) \citep{garnelo2018neural} are a new class of methods that combine the strengths of both worlds.~By taking the meta-learning framework, Neural Processes \textit{learn to learn} a stochastic process quickly from observations while experiencing multiple tasks of stochastic process modeling. Thus, in Neural Processes, unlike typical neural networks, learning a function is fast and uncertainty-aware while, unlike Gaussian processes, prediction at test time is still efficient.

An important aspect for which Neural Processes can be extended is that in many cases, certain temporal dynamics underlies in a sequence of stochastic processes.~This covers a broad range of problems from learning RL agents being exposed to increasingly more challenging tasks to modeling dynamic 3D scenes.~For instance, \cite{eslami2018neural} proposed a variant of Neural Processes, called the Generative Query Networks (GQN), to learn representation and rendering of 3D scenes.
Although this was successful in modeling static scenes like fixed objects in a room, we argue that to handle more general cases such as dynamic scenes where objects can move or interact over time, we need to explicitly incorporate a temporal transition model into Neural Processes.

In this paper, we introduce Sequential Neural Processes (SNP) to incorporate the temporal state-transition model into Neural Processes. The proposed model extends the potential of Neural Processes from modeling a stochastic process to modeling a dynamically changing sequence of stochastic processes. That is, SNP can model a (sequential) stochastic process of stochastic processes. We also propose to apply SNP for dynamic 3D scene modeling by developing the Temporal Generative Query Networks (TGQN). In experiments, we show that TGQN outperforms GQN in terms of capturing transition stochasticity, generation quality, generalization to time-horizons longer than those used during training.

Our main contributions are: We introduce Sequential Neural Processes (SNP), a meta-transfer learning framework for a sequence of stochastic processes. We realize SNP for dynamic 3D scene inference by introducing Temporal Generative Query Networks (TGQN). To our knowledge, this is the first 4D generative model that models dynamic 3D scenes. We describe the training challenge of \emph{transition-collapse} unique to  SNP modeling and resolve it by introducing the \emph{posterior-dropout} ELBO. We demonstrate the generalization capability of TGQN beyond the sequence lengths used during training. We also demonstrate meta-transfer learning and improved generation quality in contrast to Consistent Generative Query Networks~\citep{kumar2018consistent} gained from the decoupling of temporal dynamics from the scene representations.

\section{Background}
\label{sec:background}
In this section, we introduce notations and foundational concepts that underlie the design of our proposed model as well as motivating applications.

\textbf{Neural Processes.}
Neural Processes (NP) model a stochastic process mapping an input $x \in \eR^{d_x}$ to a random variable $Y \in \eR^{d_y}$. In particular, an NP is defined as a conditional latent variable model where a set of \textit{context} observations $C = (X_C,Y_C) = \{(x_i,y_i)\}_{i\in \cI(C)}$ is given to model a conditional prior on the latent variable $P(z|C)$, and the \textit{target} observations $D = (X, Y) = \{(x_i,y_i)\}_{i\in \cI(D)}$ are modeled by the observation model $p(y_i|x_i,z)$. Here, $\cI(\cS)$ stands for the set of data-point indices in a dataset $\cS$. This generative process can be written as follows:
\eq{
P(Y|X, C) = \int P(Y|X, z)P(z|C) \upd z
}
where $P(Y|X, z) = \prod_{i\in \cI(D)} P(y_i|x_i,z)$. The dataset $\{(C_i,D_i)\}_{i\in\cI_{\text{dataset}}}$ as a whole contains multiple pairs of context and target sets. Each such pair $(C,D)$ is associated with its own stochastic process from which its observations are drawn. Therefore NP flexibly models multiple tasks i.e.~stochastic processes and this results in a meta-learning framework. ~It is sometimes useful to condition the observation model directly on the context $C$ as well, i.e., $p(y_i|x_i,s_C,z)$ where $s_C = f_s(C)$ with $f_s$ a deterministic context encoder invariant to the ordering of the contexts. A similar encoder is also used for the conditional prior giving $p(z|C) = p(z|r_C)$ with $f_{r}(C)$. In this case, the observation model uses the context in two ways: a noisy latent path via $z$ and a deterministic path via $s_C$. 

The design principle underlying this modeling is to infer the target stochastic process from contexts in such a way that sampling $z$ from $P(z|C)$ corresponds to a function which is a realization of a stochastic process. 
Because the true posterior is intractable, the model is trained via variational approximation which gives the following evidence lower bound (ELBO) objective:
\eq{
\log P_\ta(Y|X, C) \geq \eE_{Q_\phi(z|C,D)}\left[ \log P_\ta(Y|X,z)\right] - \KL(Q_\phi(z|C,D) \parallel P_\ta(z|C)).
}
The ELBO is optimized using the reparameterization trick \citep{kingma2013auto}.

\textbf{Generative Query Networks.}
The Generative Query Network (GQN) can be seen as an application of the Neural Processes specifically geared towards 3D scene inference and rendering. In GQN, query $x$ corresponds to a camera viewpoint in a 3D space, and output $y$ is an image taken from the camera viewpoint. Thus, the problem in GQN is cast as: given a context set of viewpoint-image pairs, (i) to infer the representation of the full 3D space and then (ii) to generate an observation image corresponding to a given query viewpoint.

In the original GQN, the prior is conditioned also on the query viewpoint in addition to the context, i.e., $P(z|x,r_C)$, and thus results in inconsistent samples across different viewpoints when modeling uncertainty in the scene.
The Consistent GQN~\citep{kumar2018consistent} (CGQN) resolved this by removing the dependency on the query viewpoint from the prior. This resulted in $z$ to be a summary of a full 3D scene independent of the query viewpoint.~Hence, it is consistent across viewpoints and more similar to the original Neural Processes. For the remainder of the paper, we use the abbreviation \emph{GQN} for CGQN unless stated otherwise.

For inferring representations of 3D scenes, a more complex modeling of latents is needed. For this, GQN uses ConvDRAW~\citep{gregor2016towards}, an auto-regressive density estimator performing $P(z|C) = \prod_{l=1}^{L}P(z^l|z^{<l}, r_C)$ where $L$ is the number of auto-regressive rollout steps and $r_C$ is a pooled context representations $\sum_{i \in \cI(C)} f_r(x_i,y_i)$ with $f_r$ an encoding network for context.

\textbf{State-Space Models.}
State-space models (SSMs) have been one of the most popular models in modeling sequences and dynamical systems. The model is specified by a state transition model $P(z_t|z_\tmo)$ that is sometimes also conditioned on an action $a_\tmo$, and an observation model $P(y_t|z_t)$ that specifies the distribution of the (partial and noisy) observation from the latent state. Although SSMs have good properties like modularity and interpretability due to the Markovian assumption, the closed-form solution is only available for simple cases like the linear Gaussian SSMs.~Therefore, in many applications, SSMs show difficulties in capturing nonlinear non-Markovian long-term dependencies~\citep{auger2016state}. To resolve this problem, RNNs have been combined with SSMs~\citep{zheng2017state}.
In particular, the Recurrent State-Space Model (RSSM)~\citep{hafner2018learning} maintains both a deterministic RNN state $h_t$ and a stochastic latent state $z_t$ that are updated as follows:
\eq{
h_{t} &= f_\text{RNN}(h_{t-1}, z_{t-1}), \hspace{10mm} z_{t} \sim p(z_{t} | h_{t}), \hspace{10mm} y_{t} \sim p(y_{t} | h_{t}, z_{t}).  
}
Thus, in RSSM, the state transition is dependent on all the past latents $z_\lt$ and thus non-Markovian.

\section{Sequential Neural Processes}
\label{sec:snp}
In this section, we describe the proposed \emph{Sequential Neural Processes} which combines the merits of SSMs~and~Neural Processes and thus enabling it to model temporally-changing stochastic processes.

\subsection{Generative Process} 
\begin{wrapfigure}{r}{0.28\textwidth}
  \vspace{-4mm}
  \begin{center}
   \includegraphics[width=0.98\linewidth]{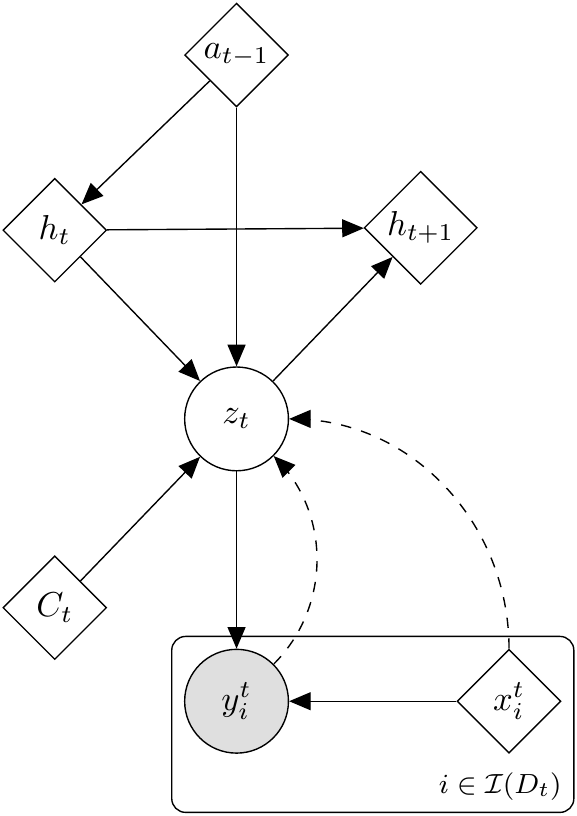}
  \end{center}
  \caption{Generative and inference (shown as dashed edges) models in TGQN.}
  \label{TGQN_fig}
\end{wrapfigure}

Consider a sequence of stochastic processes $\cP_1,\dots, \cP_T$.
At each time-step $t\in[1,T]$, for a true stochastic process $\cP_t$, consider drawing a set of context observations $C_t=\{(x_i^t, y_i^t)\}_{i\in \cI(C_t)}$ where $\cI(C_t)$ are the indices of the observations. Size of this context set may differ over time or it may even be empty. The $C_t$ are provided to the model at their respective time-steps and we want SNP to model $\cP_t$ as a distribution over a latent variable $z_t$, as modeled in NP. 

While NP models $z_t$ only using $C_t$ i.e.,~$P(z_t|C_t)$, in SNP we want to utilize the underlying temporal structure which governs the temporal change in the true stochastic processes $\cP_\tmo \ra \cP_t$. We achieve this by providing the latents of the past stochastic processes $z_\lt$ to the distribution of the current $z_t$ resulting in $P(z_t|z_\lt, C_t)$. Here $z_\lt$ may be represented as an RNN encoding. The sampled latent $z_t$ is then used to model the target observation set $D_t = (X_t, Y_t)$ through $P(Y_t|X_t,z_t)$. Like $C_t$, we assume that $D_t$ is also drawn from the true process $\cP_t$. With an abuse of notation, we use $C$, $D$, $X$, and $Y$ to bundle together the $C_t$, $D_t$, $X_t$, and $Y_t$ for all time-steps $t\in[1,T]$, e.g.,~$C=(C_1,\dots,C_T)$. With these notations, the generative process of SNP is as follows:
\eq{
P(Y,Z|X,C) = \prod_{t=1}^{T}P(Y_t|X_t,z_t)P(z_t|z_\lt,C_t)
\label{eq:gen}}
where $P(Y_t|X_t,z_t)=\prod_{i\in \cI(D_t)}P(y_i^t|x_i^t,z_t)$ and $z_0 = \texttt{null}$. The transition can also be conditioned on an action $a_\tmo$, but we omit this throughout the paper for brevity.

Although we use the RSSM version of SNP in Eqn.~\eqref{eq:gen} where the transition depends on all the past $z_\lt$, what we propose is a generic SNP class of models that is compatible with a wide range of temporal transition models including the traditional state-space model \citep{krishnan2017structured} as long as the latents do not access the previous contexts $\cC_\lt$ directly.

Some of the properties of SNPs are as follows: (i) SNPs can be seen as a \textbf{generalization of NPs} in two ways. First, if $T=1$, an SNP equals an NP. Second, if $D_t$ is empty for all $t<T$ and non-empty when $t=T$, SNP becomes an NP which uses the state transition as the (stochastic) context aggregator instead of the standard sum encoding.~It then becomes an order sensitive encoding that can in practice be dealt with the order-shuffling on the contexts $\{C_t\}$. (ii) SNPs are a \textbf{meta-transfer learning} method.~Consider, for example, a game-playing agent which, after clearing up the current stage, levels up to the next stage where more and faster enemies are placed than the previous stage. With SNP, the agent can not only meta-update the policy \textit{with only a few observations} $C_t$ from the new stage, but it can also \textit{transfer} the general trend from the past, namely, that there will be more and faster enemies in the future stages. As such, we can consider SNP to be a model combining temporal transfer-learning via $z_t$ and meta-learning via $C_t$.

\subsection{Learning and Inference}
Because a closed-form solution for learning and inference is not available for general non-linear transition and observation models, we train the model via variational approximation. For this, we approximate the true posterior with the following temporal auto-regressive factorization
\eq{
P(Z|C,D) \approx \prod_{t=1}^{T}Q_\phi(z_t|z_\lt,C,D)
\label{eq:snp_infer}
} with $z_0 = \texttt{null}$.
\cite{chung2015recurrent, fraccaro2016sequential, krishnan2017structured, hafner2018learning} provide various implementation options for the above approximation based on RNNs (forward or bi-directional) and the reparameterization-trick used. In the next section, we introduce a particular implementation of the above approximate posterior for an application to dynamic 3D-scene modeling.

With this approximate posterior, we train the model using the following evidence lower bound (ELBO): $\log P(Y|X,C) \geq \cL_\text{SNP}(\ta,\phi) = $
\eq{
 \sm{t}{T}\eE_{Q_\phi(z_t|\cV)}\left[\log P_\ta(Y_t|X_t,z_t)\right] - \eE_{Q_\phi(z_\lt|\cV)}\left[\KL(Q_\phi(z_t|z_{<t},\cV) \parallel P_\ta(z_t|z_\lt,C_t)) \right]
\label{eq:snp_elbo}
}
where $\cV = (C,D)$ and $\log P_\ta(Y_t|X_t,z_t) = \sum_{i \in \cI(D)}\log P_\ta(y_i^t|x_i^t,z_t)$. We use the reparameterization trick to compute the gradient of the objective. For the derivation of Eqn.~\eqref{eq:snp_elbo}, see Appendix~\ref{sec:snp_elbo_deriv}.

\subsection{Temporal Generative Query Networks}
\label{sec:tgqn}

Consider a room placed with an object. An agent can control the object by applying some actions such as translation or rotation. For such setups, whenever an action is applied, the scene changes and thus the viewpoint-to-image mapping of GQN learned in the past become stale because the same viewpoint now maps to a different image altogether.~Although the new scene can be learned again from scratch using new context from the new scene, an ideal model would also be able to transfer the past knowledge such as object colors as well as utilizing the action to update its belief about the new scene.~With a successful transfer, the model would adapt to the new scene with only small or no context from the new scene.

To develop this model, we propose applying SNP to extend GQN into Temporal GQN (TGQN) for modeling complex dynamic 3D scenes. In this setting, at time $t$, $C_t$ becomes the camera observations, $a_t$ the action provided to the scene objects, $z_t$ a representation of the full 3D scene, $X_t$ the camera viewpoints and $Y_t$ the images. TGQN draws upon the GQN implementation in multiple ways. We encode raw image observations and viewpoints into $C_t$ using the same encoder network and use a DRAW-like recurrent image renderer. Unlike GQN, to capture the transitions, we introduce the Temporal-ConvDRAW (T-ConvDRAW) where we condition $z_t^l$ on the past $z_\lt$ via a concatenation of $(C_t, h_t, a_t)$. That is, $P(z_t|z_\lt,C_t) = \prod_{l=1}^{L} P(z_t^l|z_t^{<l},z_\lt,C_t)$. Taking an RSSM approach \citep{hafner2018learning}, $h_t$ is transitioned using a ConvLSTM \citep{xingjian2015convolutional}. (See Fig.~\ref{TGQN_fig}). In inference, to realize the distribution in Equation~\eqref{eq:snp_infer}, $C_t \cup D_t$ is provided like in GQN (see Appendix \ref{ax:tgqn-nn}).

\subsection{Posterior Dropout for Mitigating Transition Collapse}
\label{sec:posterior_dropout}
A novel part of SNP model is the use of the state transition $P(z_t|z_\lt,C_t)$ which is not only conditioned on the past latents $z_\lt$ but also on the context $C_t$.~While this makes our model perform the meta-transfer learning, we found that it creates a tendency to ignore the context $C_t$ in the transition model. It seems that the problem lies in the KL term in Eqn.~\eqref{eq:snp_elbo} which drives the training of the transition $p_\ta(z_t|z_{<t},C_t)$. We note that the two distributions $q_\phi$ and $p_\ta$ are conditioned on the previous latents $z_\lt$ which are sampled by providing all the available information $C$ and $D$. This produces a rich posterior with low uncertainty that makes good reconstructions via the decoder. While this is desirable modeling in general, we found that in practice it can make the KL collapse as the transition relies more on $z_{<t}$ while ignoring $C_t$. 

This is a similar but not the same problem as the posterior collapsing~\citep{bowman2015generating} because in our case the cause of the collapse is not an expressive decoder (e.g., auto-regressive), but a conditional prior which is already provided rich information about the sequence of tasks from one path via $z_\lt$ and thus open a possibility to ignore the other path $C_t$. We call this the \textit{transition collapse} problem. 

To resolve this, we need a way to (i) limit the information available in $z_\lt$ to incentivize the use of $C_t$ information when available while (ii) maintaining the high quality of the reconstructions. We introduce the \textit{posterior-dropout} ELBO where we randomly choose a subset of time-steps $\cT \subseteq [1,T]$. For these time-steps, the $z_t$ are sampled using the prior transition $p_\ta$. For the remaining time-steps in $\bar{\cT} \equiv [1,T] \setminus \cT$, the $z_t$ are sampled using the posterior transition $q_\phi$. This leads to the following approximate posterior:
\begin{align}
    \tilde{Q}(Z) &= \prod_{t \in \cT}P_\ta(z_t|z_{<t},C_t)\prod_{t \in \bar{\cT}} Q_\phi(z_{t}|z_{<t},C,D)
\end{align}
Such a posterior limits the information contained in the past latents $z_\lt$ and encourages $p_\ta$ to use the context $C_t$ for reducing the KL term. Furthermore, we reconstruct images only for time-steps $t \in \bar{\cT}$ using latents sampled from $q_\phi$. This is because reconstructing the observations at those time-steps that use prior transitions does not satisfy the principle of auto-encoding, i.e.,~it then tries to reconstruct an observation that is not provided to the encoder and, not surprisingly, would result in blurry reconstructions and poorly disentangled latent space. 
Therefore, the posterior-dropout ELBO becomes:
$\eE_{\tilde{\cT}}\log P(Y_{\tilde{\cT}}|X,C) \geq \cL_{\text{PD}}(\ta,\phi) =$
\eq{
  \eE_{\tilde{\cT}}\left[\eE_{Z\sim \tilde{\cQ}}\left[ \sum_{t\in\tilde{\cT}}\left[ \log P_\ta(Y_t|X_t,z_t) - \KL\left( Q_\phi(z_t|z_{<t},C,D) \parallel P_\ta(z_t|z_{<t},C_t) \right) \right]\right]\right]\label{eq:aux_elbo}
}
 Combining~\eqref{eq:snp_elbo} and~\eqref{eq:aux_elbo}, we take the complete maximization objective as $\mathcal{L}_{\text{SNP}} + \al \mathcal{L}_{\text{PD}}$ with $\al$ an optional hyper-parameter. In experiments, we simply set $\alpha=0$ at the start of the training and set $\alpha=1$ when the reconstruction loss had saturated (see Appendix \ref{ax:hyperparam}). For derivation of Eqn.~\eqref{eq:aux_elbo}, see Appendix~\ref{sec:aux_elbo_deriv}.

\section{Related Works}
\label{sec:related_work}

Modeling flexible stochastic processes with neural networks has seen significant interest in recent times catalyzed by its close connection to meta-learning. Conditional Neural Processes (CNP)~\citep{garnelo2018conditional} is a precursor to Neural Processes ~\citep{garnelo2018neural} which models the stochastic process \emph{without} an explicit global latent. Without it, the sampled outputs at different query inputs are uncorrelated given the context. This is addressed by NP by introducing an explicit latent path. A discussion on NP, GQN \citep{eslami2018neural} and CGQN \citep{kumar2018consistent} has been presented in Sec.~\ref{sec:background}. To improve the NP modeling further, one line of work pursues the problem of under-fitting of the meta-learned function on the context. To resolve this, attention on the relevant context points at query time is shown to be beneficial in ANP \citep{kim2019attentive}. \cite{rosenbaum2018learning} apply GQN to more complex 3D maps (such as in Minecraft) by performing patch-wise attention on the context images. 

In the domain of SSMs, Deep Kalman Filters \citep{krishnan2017structured} and DVBF \citep{karl2016deep} consist of Markovian state transition models for the hidden latents and an emission model for the observations. But instead of a Markovian latent structure, VRNN \citep{chung2015recurrent} and SRNN \citep{fraccaro2016sequential} introduce skip-connections to the past latents making roll-out auto-regressive. \cite{zheng2017state} and \cite{hafner2018learning} propose Recurrent State-Space Models which also takes advantage of the RNNs to model long-term non-linear dependencies. Other variants and inference approximations have been explored by \cite{buesing2018learning}, \cite{fraccaro2017disentangled}, \cite{elef2017ident}, \cite{goyal2017z} and \cite{krishnan2017structured}. To further model the long-term nonlinear dependencies, \cite{gemici2017generative} and \cite{fraccaro2018generative} attach a memory to the transition models. Mitigating transition-collapse through posterior-dropout broadly tries to bridge the gap between what the transition model sees during training and the test time. This intuition is related to \emph{scheduled sampling} introduced by~\cite{bengio2015scheduled} which mitigates the teacher-forcing problem.

\section{Experiments}
\label{sec:expr}
We evaluate SNP on a toy regression task, and 2D and 3D scene modeling tasks. We use NP and CGQN as the baselines. We note that these baselines, unlike our model, directly access all the context data points observed in the past at every time-step of an episode and thus result in a strong baseline.

\subsection{Regression}
We generate a dataset consisting of sequences of functions. Each function is drawn from a Gaussian process with squared-exponential kernels. For temporal dynamics between consecutive functions in the sequence, we gradually change the kernel hyper-parameters with an update function and add a small Gaussian noise for stochasticity. For more details on the data generation, see Appendix \ref{ax:1d_gp}. 

We explore three sub-tasks with different context regimes. In task (a), we are interested in how the transition model generalizes over the time steps. Therefore, we provide context points only in the first 10 time-steps out of 20. In task (b), we provide the context intermittently on randomly chosen 10 time steps out of 20. Our goal is to see how the model incorporates the new context information and updates its belief about the time-evolving function. In (a) and (b), the number of revealed points are randomly picked between 5 and 50 for each time-step chosen for showing the context. On the contrary, in task (c), we shrink this context size to 1 and provide it in 45 randomly chosen time-steps out of 50. Our goal is to test how such highly partial observations can be accumulated and retained over the long-term. The models were trained in these settings before performing validation. In Appendix \ref{1d_model}, we describe the architectures of SNP and the baseline NP for the 1D regression setting.

We present our quantitative results in Fig.~\ref{fig:cs_cc_mo_eval}. We report the target NLL on a held-out set of 1600 episodes computed by sampling the latents conditioned on the context as in \cite{kim2019attentive}. In task (a), in the absence of context for $t\in[11,20]$ we expect the transition noise to accumulate for any model since the underlying true dynamics are also noisy. We note that in contrast to NP, SNP shows less degradation in prediction accuracy.  In task (b) and (c) as well, the proposed SNP outperforms the NP baseline. In fact, SNP's accuracy improves with accumulating context while NP's accuracy deteriorates with time. This is particularly interesting because NP is allowed to access the past context directly whereas SNP is not. This demonstrates a more effective transfer of past knowledge in contrast to the baseline. More qualitative results are provided in Appendix  \ref{ax:unc_demo} (Fig. \ref{fig:reg_gen_vz}). 
PD was not particularly crucial for training success on the 1D regression tasks (see Fig.~\ref{fig:cs_cc_mo_eval}). Fig.~\ref{fig:1d_sample} compares the sampled functions.

\subsection{2D and 3D Dynamic Scene Inference}
We subject our model to the following 2D and 3D visual scene environments. The \emph{2D environments} consist of a white canvas having two \textit{moving} objects. Objects are picked with a random shape and color which, to test stochastic transition, may randomly be changed once in any episode with a fixed rule e.g.,~red $\leftrightarrow$ magenta or blue $\leftrightarrow$ cyan. When two objects overlap, one covers the other based on a fixed rule (See Appendix~\ref{ax:2d_colorshapes}). Given a 2D viewpoint, the agent can observe a $64\times64$-sized cropped portion of the canvas around it. The \emph{3D environments} consist of movable object(s) inside a walled-enclosure. The camera is always placed on a circle facing the center of the arena. Based on the camera's angular position $u$, the query viewpoint is a vector $(\cos{u}, \sin{u},u)$. We test the following two 3D environments: \begin{enumerate*}[label=\itshape\alph*\upshape)]
\item \emph{Color Cube Environment} contains a cube with different colors on each face. The cube moves or rotates at each time-step based on the translation actions (Left, Right, Up, Down) and the rotation actions (Anti-clockwise, Clockwise)
\item \emph{Multi-Object Environment:} The arena contains a randomly colored sphere, a cylinder and a cube with translation actions given to them (see Appendix \ref{ax:mujoco}).
\end{enumerate*} The action at each time-step is chosen uniformly. The 3D datasets have two versions: \textit{deterministic} and \textit{jittery}. In the former, each action has a deterministic effect on the objects. In the jittery version, a small Gaussian jitter is added to the object motion after the action is executed. The purpose of these two versions is described next.

\textbf{Context Regimes.} We explore two kinds of context regimes: \emph{prediction} and \emph{tracking}. In the \emph{prediction} regime, we evaluate the model's ability to predict future time-steps without any assistance from the context. So we provide up to 4 observations in each of the first 5 time-steps and let the model predict the remaining time-steps (guided only by the actions in the 3D tasks). We also predict beyond the training sequence length ($T=10$) to test the generalization capability. This regime is used with the 2D and the deterministic 3D datasets. In the \emph{tracking} regime, we seek to demonstrate how the model can transfer past knowledge while also meta-learning the process from the partial observations of the current time-step. We, therefore, provide only up to 2 observations at every time-step of the roll-out of length $T=20$. We test this regime with the 2D and the jittery 3D datasets since, in these settings, the model would keep finding new knowledge in every observation.

\textbf{Baseline and Performance Metrics.} We compare TGQN to GQN as baseline. Since GQN's original design does not consume actions, we concatenate the camera viewpoint and the RNN encoding of the action sequence up to that time-step to form the GQN query. In the action-less environments, the query is the camera viewpoint concatenated with the normalized $t$ (see Appendix~\ref{ax:gqn_baseline}). We report the NLL of the entire roll out $-\log P(Y|X,C)$ estimated using 40 samples of $Z$ from $Q(Z|C,D)$. To report the time-step wise generation quality, we compute the pixel MSE per target image averaged over 40 generated samples using the prior $P(Z|C)$. 
\begin{figure}
\begin{minipage}[t]{0.49\linewidth}
\vspace{0.4mm}
\hspace{-4.5mm}
\centering
\includegraphics[width=1.05\linewidth]{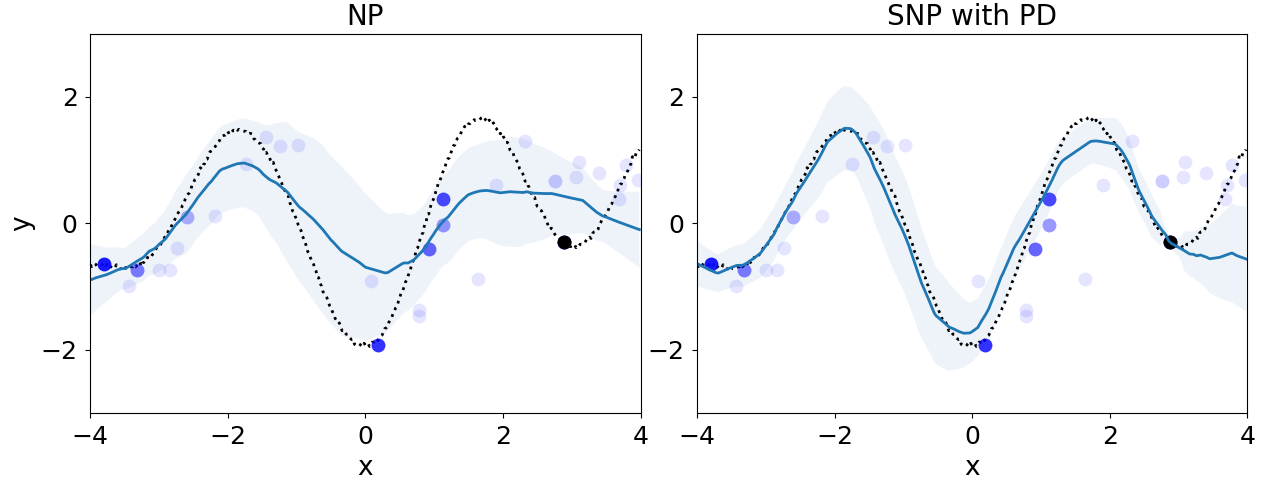}
\captionof{figure}{Sample prediction in 1D regression task (c) at $t=33$.
\textit{Blue dots:} Past context. \textit{Black dots:} Current context. \textit{Black dotted line:} True function. \textit{Blue line:} Prediction. \textit{Blue shaded region:} Prediction uncertainty.}
\label{fig:1d_sample}
\end{minipage}
\hfill
\begin{minipage}[t]{0.48\linewidth}
\vspace{0pt}
\centering
\small
\tabcolsep=0.09cm
\begin{tabular}[t]{@{}lllrrr@{}}
\toprule
Dataset      & Regime    & $T$ & \multicolumn{1}{l}{GQN} & \multicolumn{2}{c}{TGQN}                                     \\ \cmidrule(l){5-6} 
             &            &     & \multicolumn{1}{l}{}    & \multicolumn{1}{l}{no PD} & \multicolumn{1}{l}{PD} \\ \midrule
Color Shapes & Predict & 20  &          5348               &      489                         &             564                \\
Color Cube (\textit{Det.})   & Predict & 10  &   380                       &         221                        &     226                         \\
Multi-Object (\textit{Det.}) & Predict & 10  &   844                       &        346                         &   357                           \\ \midrule
Color Shapes & Track & 20  &     5285           &     482        & 513 \\
Color Cube (\textit{Jit.})  & Track   & 20  &     783                   &                       153          &   156                           \\
Multi-Object (\textit{Jit.}) & Track   & 20  &  1777                    &  450                            &  475                         \\ \bottomrule
\end{tabular}
\captionof{table}{Negative $\log p(Y|X,C)$ estimated using importance-sampling from posterior with $K=40$.}
\label{tab:nll}
\end{minipage}
\end{figure}
\begin{figure}[t]
    \centering
    \begin{subfigure}[t]{0.21\textwidth}
        \centering
        \includegraphics[width=1.0\linewidth]{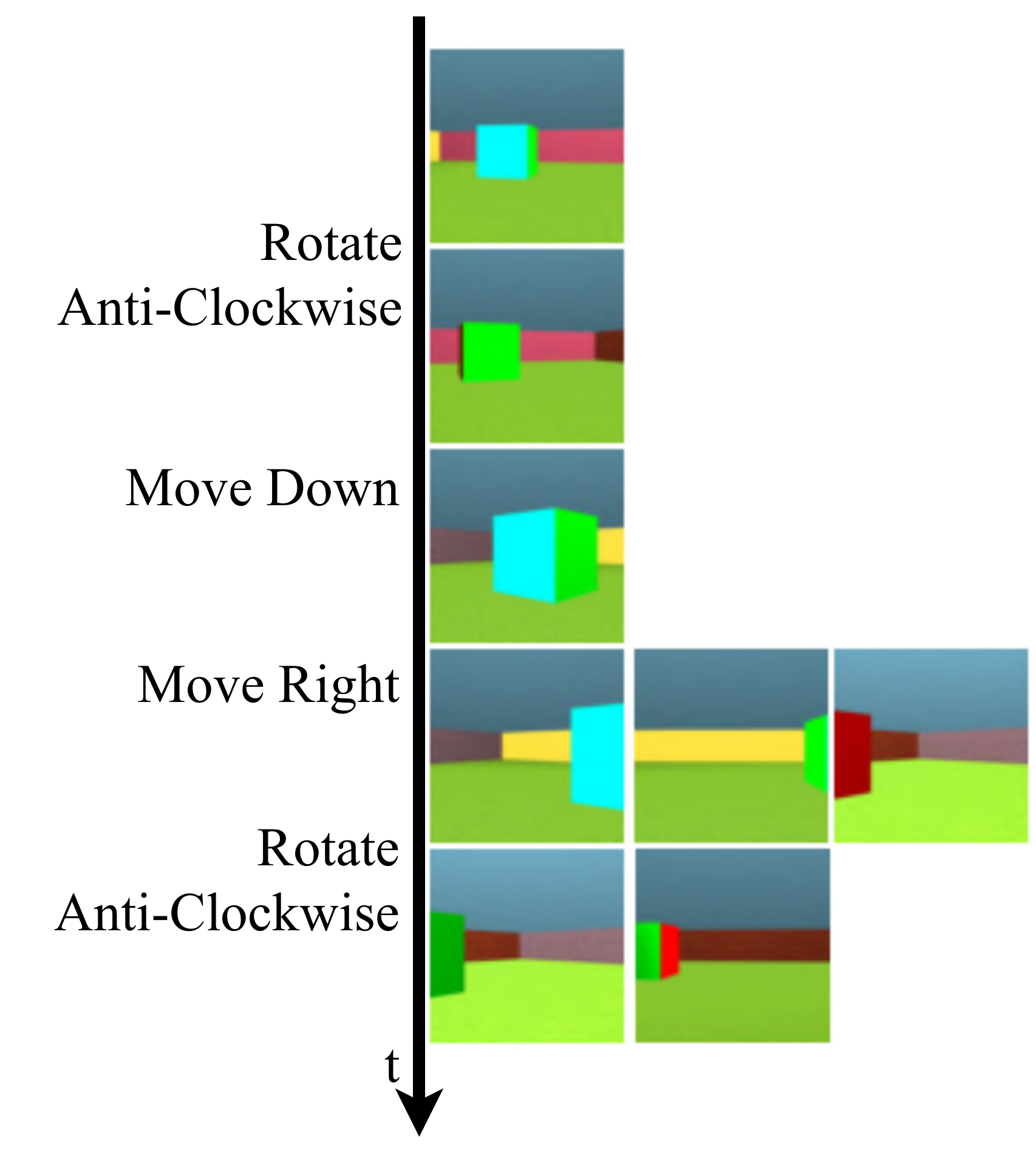}
    \caption{Context Set}
    \end{subfigure}
    \begin{subfigure}[t]{0.78\textwidth}
        \centering
        \includegraphics[width=1.0\linewidth]{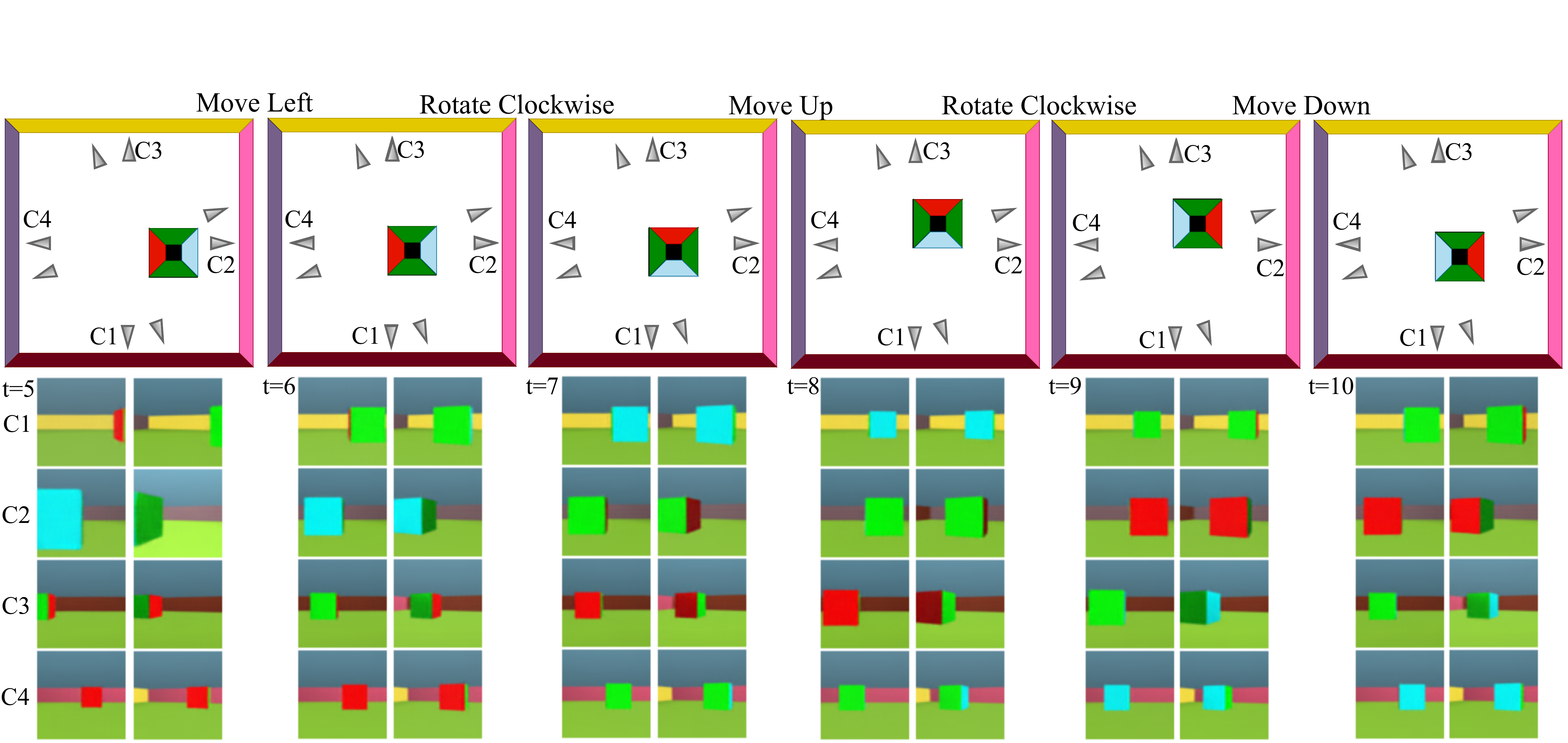}
    \caption{Generation Roll-Out}
    \end{subfigure}
    \caption{TGQN demonstration in Color-Cube Environment. \textit{Left:} The contexts and actions provided in $t<5$. \textit{Top Right:} Scene maps showing the queried camera locations and the true cube and the wall colors. \textit{Bottom Right:} TGQN predictions in $5\leq t \leq 10$. }
    \label{fig:cc_illustration}
\end{figure}
\begin{figure}[t]
    \centering
    \begin{subfigure}{0.334\textwidth}
        \centering
        \includegraphics[width=1.0\linewidth]{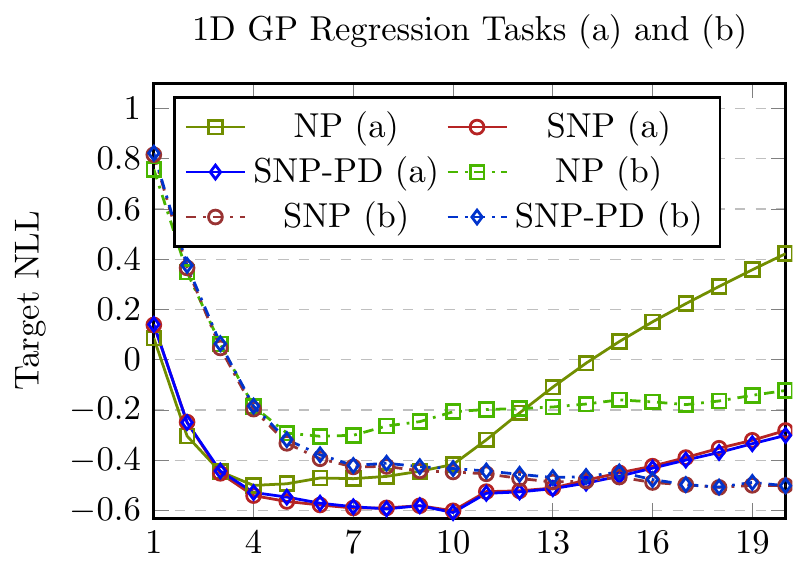}
    \end{subfigure}
    \hfill
    \begin{subfigure}{0.334\textwidth}
        \centering
        \includegraphics[width=1.0\linewidth]{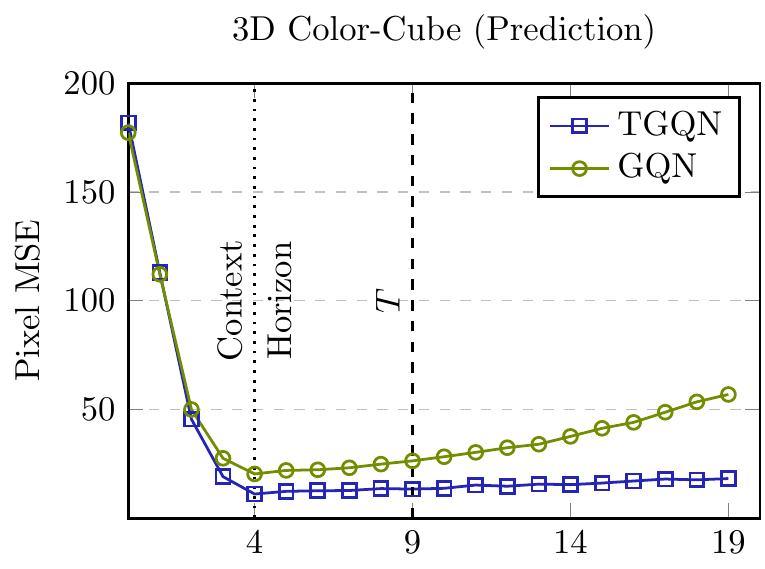}
    \end{subfigure}
   \begin{subfigure}{0.32\textwidth}
        \centering
        \includegraphics[width=1.0\linewidth]{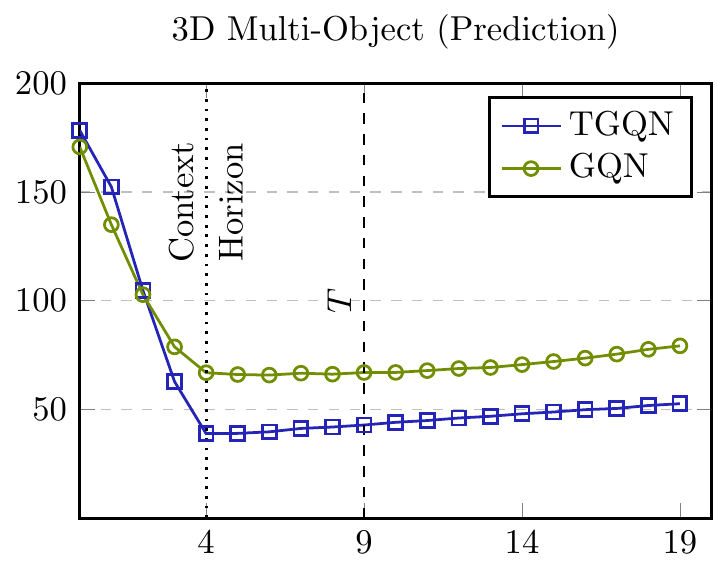}
    \end{subfigure}
\\
    \begin{subfigure}{0.334\textwidth}
        \centering
        \includegraphics[width=1.0\linewidth]{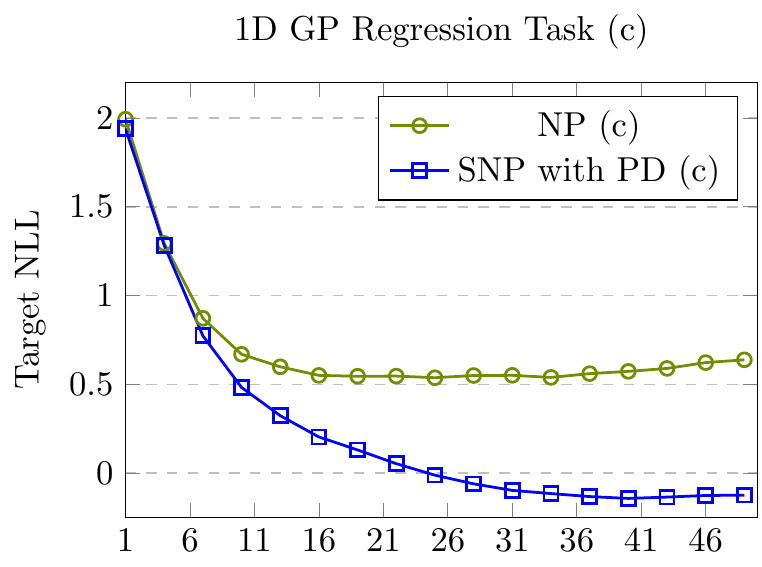}
    \end{subfigure}
    \hfill
    \begin{subfigure}{0.334\textwidth}
        \centering
        \includegraphics[width=1.0\linewidth]{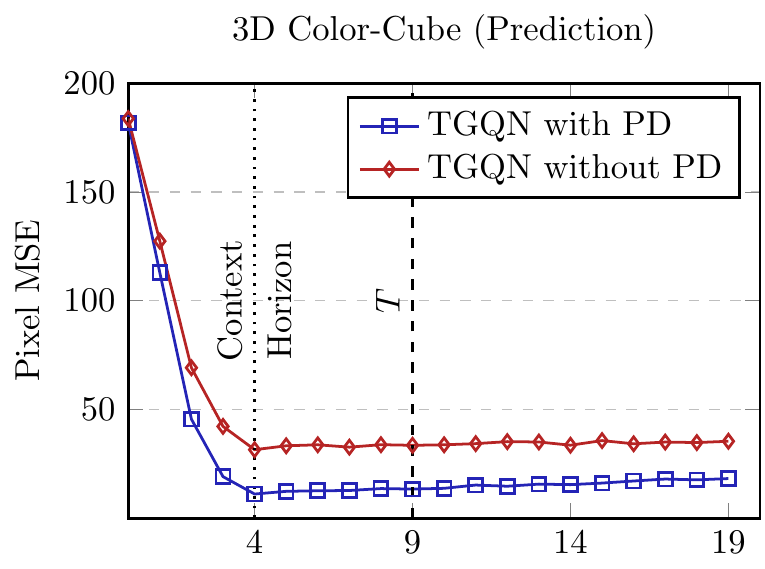}
    \end{subfigure}
   \begin{subfigure}{0.32\textwidth}
        \centering
        \includegraphics[width=1.0\linewidth]{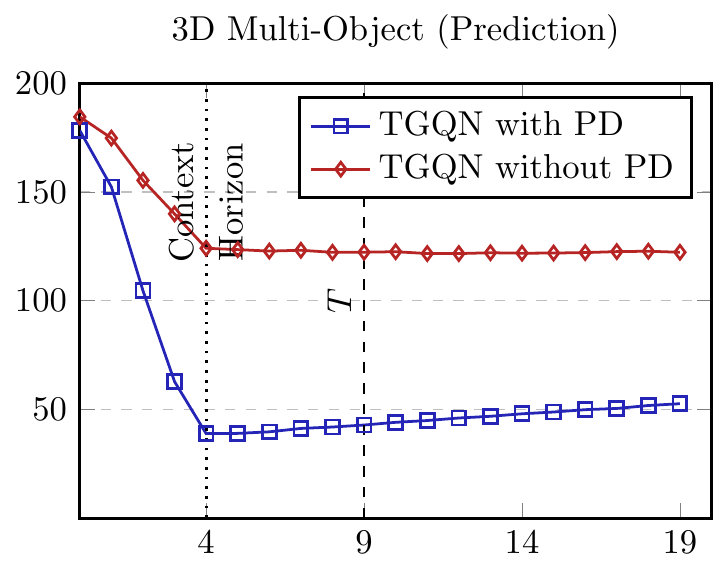}
    \end{subfigure}
    \\
    \begin{subfigure}{0.334\textwidth}
        \centering
        \includegraphics[width=1.0\linewidth]{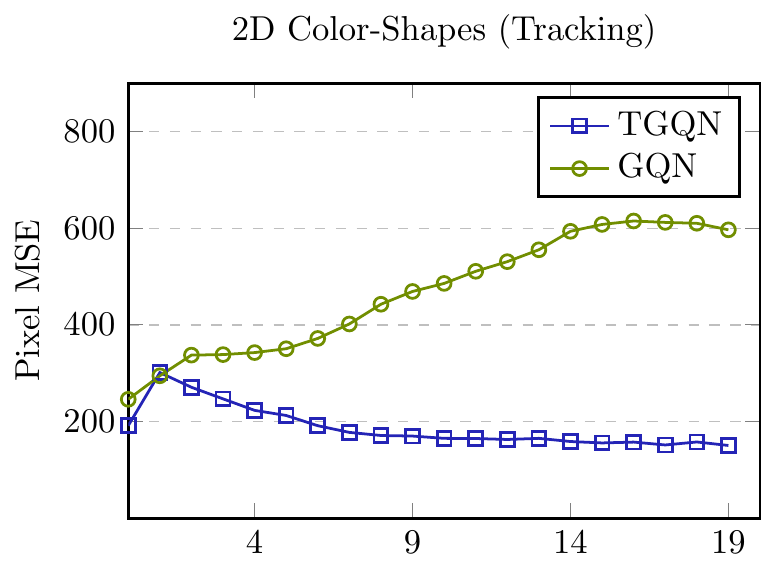}
    \end{subfigure}
    \hfill
    \begin{subfigure}{0.315\textwidth}
        \centering
        \includegraphics[width=1.0\linewidth]{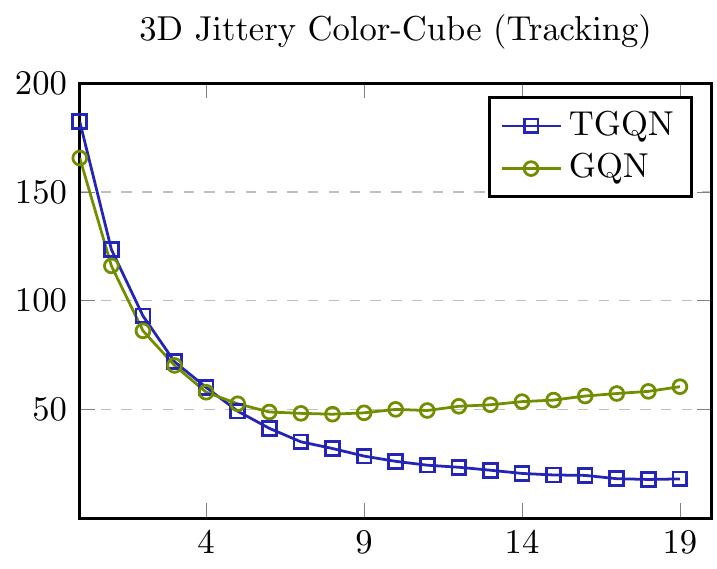}
    \end{subfigure}
    \begin{subfigure}{0.32\textwidth}
        \centering
        \includegraphics[width=1.0\linewidth]{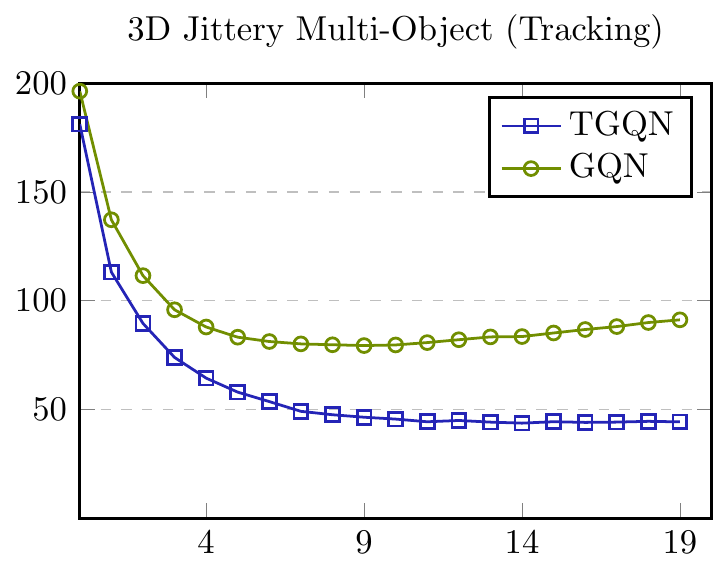}
    \end{subfigure}
    \\
    \begin{subfigure}{0.334\textwidth}
        \centering
        \includegraphics[width=1.0\linewidth]{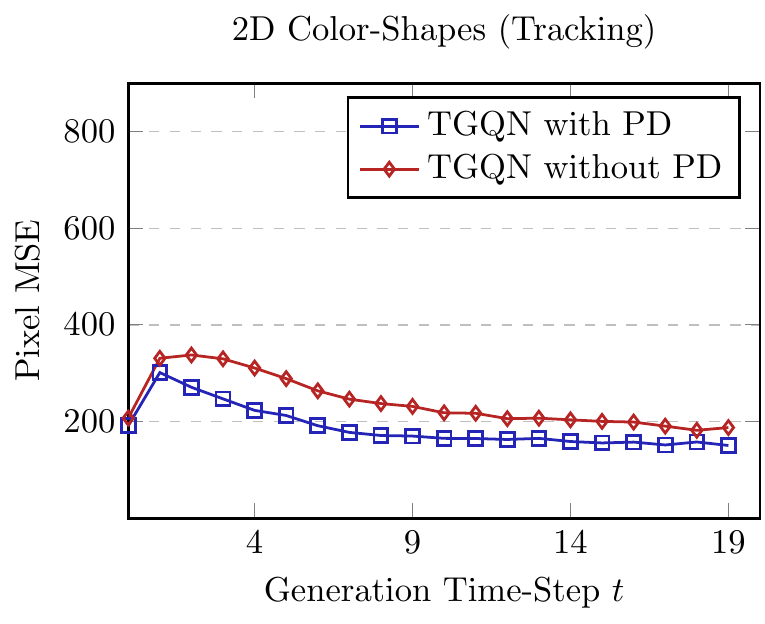}
    \end{subfigure}
    \hfill
    \begin{subfigure}{0.315\textwidth}
        \centering
        \includegraphics[width=1.0\linewidth]{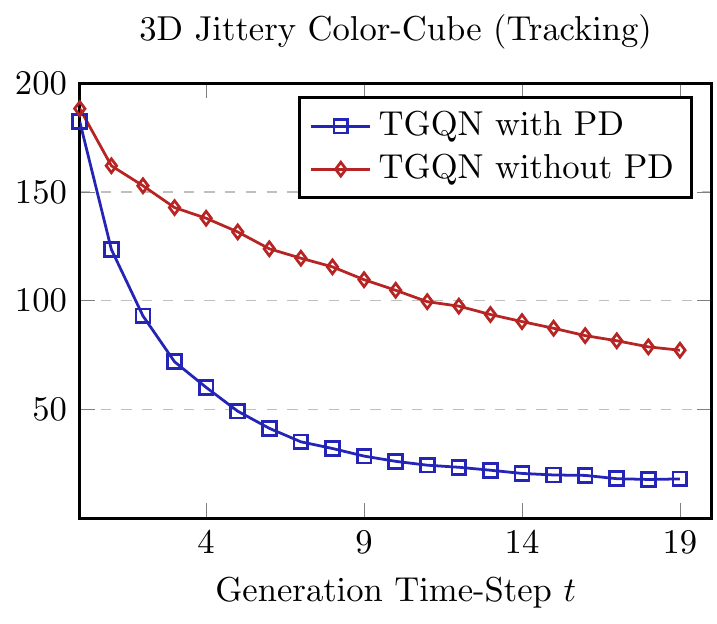}
    \end{subfigure}
    \begin{subfigure}{0.32\textwidth}
        \centering
        \includegraphics[width=1.0\linewidth]{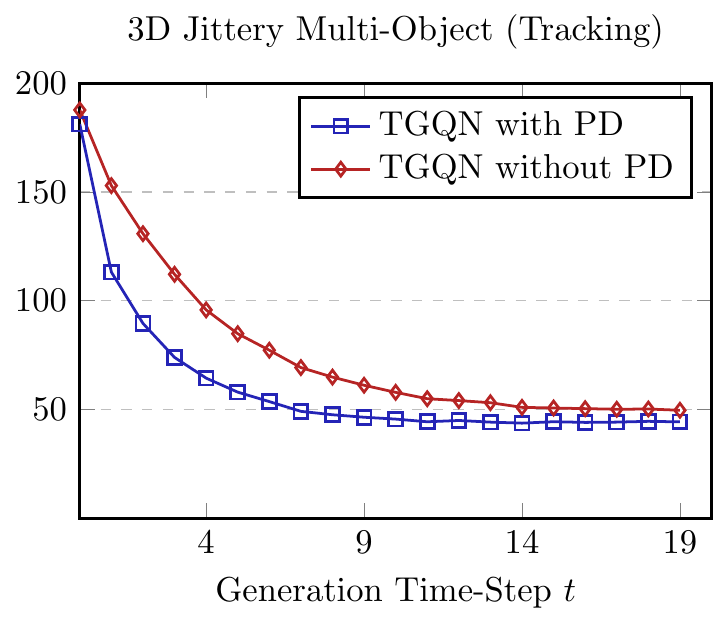}
    \end{subfigure}
    \caption{Comparison of generations of SNP with NP or GQN and comparison between SNP with and without posterior-dropout (PD). The latents are rolled-out from the prior conditioned on the context. For 1D regression, we report target NLL. For 2D and 3D settings, we report pixel MSE per image generated at each time-step.}
    \label{fig:cs_cc_mo_eval}
\end{figure}

\textbf{Quantitative Analysis.} In Table~\ref{tab:nll} and Fig.~\ref{fig:cs_cc_mo_eval}, we compare TGQN trained with posterior dropout (PD) versus GQN and versus TGQN trained without PD. TGQN outperforms GQN in all environments in both NLL and pixel MSE. In terms of image generation quality in the \emph{prediction} regime, the pixel MSE gap is sustained even beyond the training horizon. In \emph{tracking} regime, TGQN with PD converges in the fewest time-steps of observing the contexts. While TGQN continually improves by observing contexts over time, GQN's performance starts to deteriorate after a certain point. This is interesting since GQN can directly access all the past observations. This demonstrates TGQN's better temporal modeling and transfer of past knowledge. In general, the use of PD improves generation quality in all the explored cases. However, we note that the NLL of TGQN with PD is slightly higher than TGQN without PD. This is reasonable because TGQN with PD does not ignore $C_t$ when the past scene modeling in $z_{<t}$ is incorrect. This means that the model must carry extra modeling power to temporarily model the incorrect scene until more observations are available and then remodel the correct scene latent. This explains the tendency towards a slightly higher NLL.

\begin{figure}[t]
    \centering
    \includegraphics[width=1.0\linewidth]{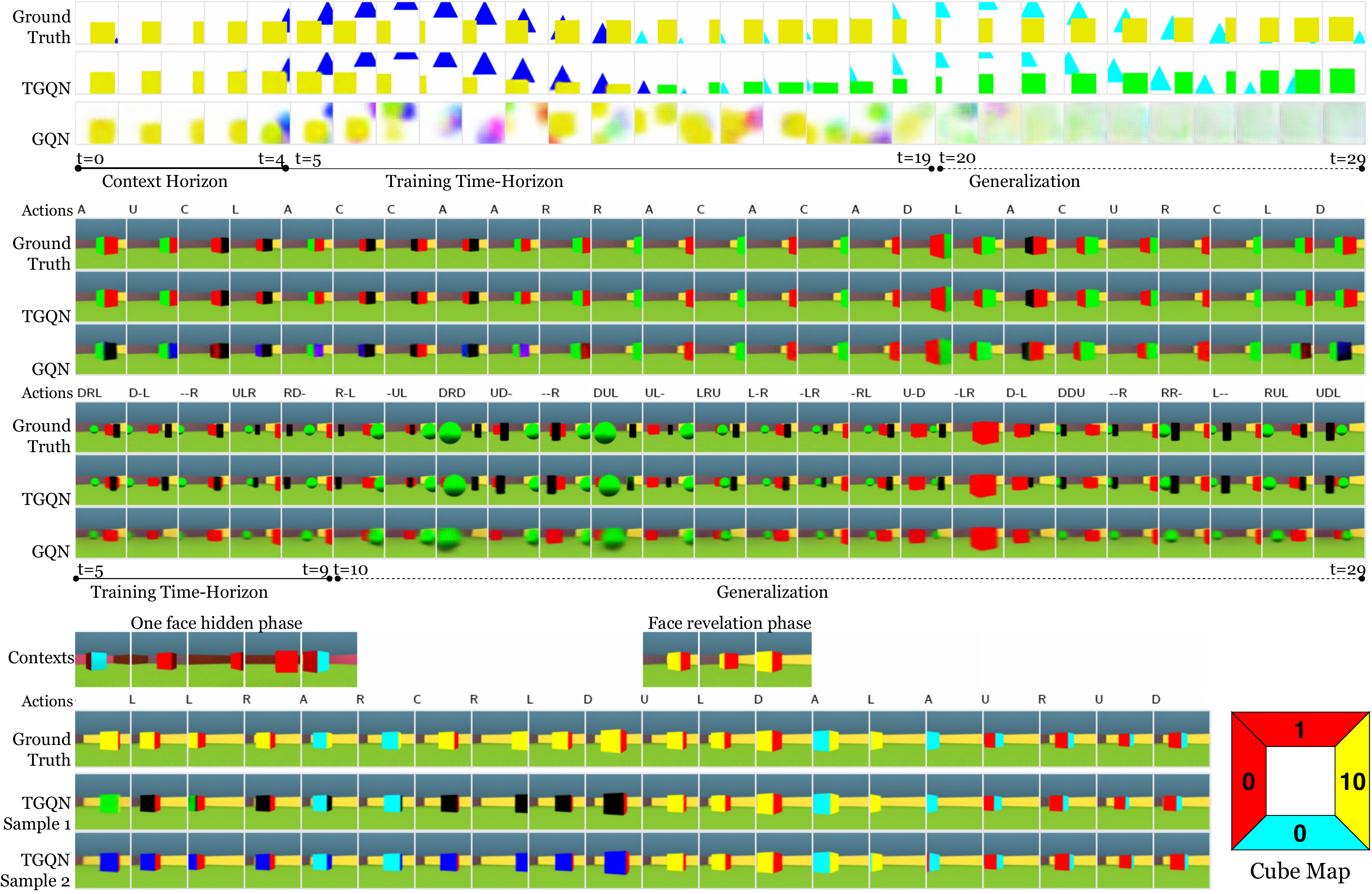}
    \caption{Qualitative comparison of TGQN with GQN (more in Appendix~\ref{ax:unc_demo} and~\ref{sec:comp_examples}). \textit{Top:} Prediction and generalization in 2D and deterministic 3D  tasks. \textit{Bottom:} Uncertainty modeling and meta-transfer learning in 3D jittery color-cube data set. The cube map shows the true face colors and the time-step at which it is revealed.}
    \label{fig:generalization}
\end{figure}

\textbf{Qualitative Analysis.} In Fig.~\ref{fig:cc_illustration}, we show a demonstration of TGQN's predictions for the Color Cube task. In Fig.~\ref{fig:generalization}, we qualitatively show the TGQN generations compared against the true images and the GQN generations. We infer the following from the figure.
\begin{enumerate*}[label=\itshape\alph*\upshape)]
\item The dynamics modeled using $p_\ta(z_t|z_{<t},C_t)$, can be used to sample long possible futures.~This differentiates our modeling from the baselines where a single latent $z$ must compress all the indefinite future possibilities. In the 2D task, TGQN keeps generating plausible shape, motion and color changes. GQN fails here because the sampled $z$ does not contain information beyond $t=20$, its training sequence-length. 
\item In the Color Cube and the Multi-Object tasks, we observe that TGQN keeps executing the correct object transitions.  In contrast, GQN is susceptible to forgetting the face colors in longer-term generations. Although GQN can generate object positions correctly, this can be credited to the RNN that encodes the action sequence into the query. (Note that this RNN action-encoding is what we additionally endow to the vanilla GQN to make a strong baseline.) However, since this RNN is deterministic, this modeling would fail to capture stochasticity in the transitions.
\item GQN models the whole roll-out in a single latent. It is therefore limited in its capacity in modeling finer details of the image. We see this through the poorer reconstruction and generation quality in the 3D tasks. 
\item TGQN can model uncertainty and perform \emph{meta-transfer learning}. We test this in the jittery color-cube task by avoiding revealing the yellow face in the early context and then revealing it at a later time-step. When the yellow face is unseen, TGQN samples a face color from the true distribution. Upon seeing the face, it updates its belief and makes the correct color while still remembering the face colors seen earlier.
\end{enumerate*}

\section{Conclusion}
We introduced SNP, a generic modeling framework for meta-learning temporally-evolving stochastic processes.
We showed that this allows for richer scene representations evidenced by the improved generation quality that can generalize to longer time-horizons in contrast to NP and GQN while also performing meta-transfer learning. We resolved the problem of transition collapse in training SNP using posterior dropout. This work leaves multiple avenues for improvement. NPs are susceptible to under-fitting \citep{kim2019attentive} and it may also be the case with SNP. It would be interesting to see how the efficiency on the number of observations needed to meta-learn new information could be improved. It would also be interesting to see if an SNP-augmented RL agent can perform better in meta-RL settings than the one without.

\newpage
\subsubsection*{Acknowledgments}
This work was supported by Electronics and Telecommunications Research Institute (ETRI) grant funded by the Korean government. [19ZH1100, Distributed Intelligence Core Technology of Hyper-Connected Space]. SA thanks to Kakao Brain, Center for Super Intelligence (CSI), and Element AI for their support. JY thanks to Kakao Brain and SAP for their support.

\bibliographystyle{icml2019}
\bibliography{refs}

\begin{thebibliography}{29}
\providecommand{\natexlab}[1]{#1}
\providecommand{\url}[1]{\texttt{#1}}
\expandafter\ifx\csname urlstyle\endcsname\relax
  \providecommand{\doi}[1]{doi: #1}\else
  \providecommand{\doi}{doi: \begingroup \urlstyle{rm}\Url}\fi

\bibitem[Abadi et~al.(2016)Abadi, Barham, Chen, Chen, Davis, Dean, Devin,
  Ghemawat, Irving, Isard, et~al.]{abadi2016tensorflow}
Abadi, M., Barham, P., Chen, J., Chen, Z., Davis, A., Dean, J., Devin, M.,
  Ghemawat, S., Irving, G., Isard, M., et~al.
\newblock Tensorflow: A system for large-scale machine learning.
\newblock In \emph{12th $\{$USENIX$\}$ Symposium on Operating Systems Design
  and Implementation ($\{$OSDI$\}$ 16)}, pp.\  265--283, 2016.

\bibitem[Auger-M{\'e}th{\'e} et~al.(2016)Auger-M{\'e}th{\'e}, Field, Albertsen,
  Derocher, Lewis, Jonsen, and Flemming]{auger2016state}
Auger-M{\'e}th{\'e}, M., Field, C., Albertsen, C.~M., Derocher, A.~E., Lewis,
  M.~A., Jonsen, I.~D., and Flemming, J.~M.
\newblock State-space models’ dirty little secrets: even simple linear
  gaussian models can have estimation problems.
\newblock \emph{Scientific reports}, 6:\penalty0 26677, 2016.

\bibitem[Bengio et~al.(2015)Bengio, Vinyals, Jaitly, and
  Shazeer]{bengio2015scheduled}
Bengio, S., Vinyals, O., Jaitly, N., and Shazeer, N.
\newblock Scheduled sampling for sequence prediction with recurrent neural
  networks.
\newblock In \emph{Advances in Neural Information Processing Systems}, pp.\
  1171--1179, 2015.

\bibitem[Bowman et~al.(2015)Bowman, Vilnis, Vinyals, Dai, Jozefowicz, and
  Bengio]{bowman2015generating}
Bowman, S.~R., Vilnis, L., Vinyals, O., Dai, A.~M., Jozefowicz, R., and Bengio,
  S.
\newblock Generating sentences from a continuous space.
\newblock \emph{arXiv preprint arXiv:1511.06349}, 2015.

\bibitem[Brockman et~al.(2016)Brockman, Cheung, Pettersson, Schneider,
  Schulman, Tang, and Zaremba]{brockman2016openai}
Brockman, G., Cheung, V., Pettersson, L., Schneider, J., Schulman, J., Tang,
  J., and Zaremba, W.
\newblock Openai gym.
\newblock \emph{arXiv preprint arXiv:1606.01540}, 2016.

\bibitem[Buesing et~al.(2018)Buesing, Weber, Racaniere, Eslami, Rezende,
  Reichert, Viola, Besse, Gregor, Hassabis, et~al.]{buesing2018learning}
Buesing, L., Weber, T., Racaniere, S., Eslami, S., Rezende, D., Reichert,
  D.~P., Viola, F., Besse, F., Gregor, K., Hassabis, D., et~al.
\newblock Learning and querying fast generative models for reinforcement
  learning.
\newblock \emph{arXiv preprint arXiv:1802.03006}, 2018.

\bibitem[Chung et~al.(2015)Chung, Kastner, Dinh, Goel, Courville, and
  Bengio]{chung2015recurrent}
Chung, J., Kastner, K., Dinh, L., Goel, K., Courville, A.~C., and Bengio, Y.
\newblock A recurrent latent variable model for sequential data.
\newblock In \emph{Advances in neural information processing systems}, pp.\
  2980--2988, 2015.

\bibitem[Eleftheriadis et~al.(2017)Eleftheriadis, Nicholson, Deisenroth, and
  Hensman]{elef2017ident}
Eleftheriadis, S., Nicholson, T., Deisenroth, M., and Hensman, J.
\newblock Identification of gaussian process state space models.
\newblock In \emph{Advances in neural information processing systems}, pp.\
  5309--5319, 2017.

\bibitem[Eslami et~al.(2018)Eslami, Rezende, Besse, Viola, Morcos, Garnelo,
  Ruderman, Rusu, Danihelka, Gregor, et~al.]{eslami2018neural}
Eslami, S.~A., Rezende, D.~J., Besse, F., Viola, F., Morcos, A.~S., Garnelo,
  M., Ruderman, A., Rusu, A.~A., Danihelka, I., Gregor, K., et~al.
\newblock Neural scene representation and rendering.
\newblock \emph{Science}, 360\penalty0 (6394):\penalty0 1204--1210, 2018.

\bibitem[Fraccaro et~al.(2016)Fraccaro, S{\o}nderby, Paquet, and
  Winther]{fraccaro2016sequential}
Fraccaro, M., S{\o}nderby, S.~K., Paquet, U., and Winther, O.
\newblock Sequential neural models with stochastic layers.
\newblock In \emph{Advances in neural information processing systems}, pp.\
  2199--2207, 2016.

\bibitem[Fraccaro et~al.(2017)Fraccaro, Kamronn, Paquet, and
  Winther]{fraccaro2017disentangled}
Fraccaro, M., Kamronn, S., Paquet, U., and Winther, O.
\newblock A disentangled recognition and nonlinear dynamics model for
  unsupervised learning.
\newblock In \emph{Advances in Neural Information Processing Systems}, pp.\
  3601--3610, 2017.

\bibitem[Fraccaro et~al.(2018)Fraccaro, Rezende, Zwols, Pritzel, Eslami, and
  Viola]{fraccaro2018generative}
Fraccaro, M., Rezende, D., Zwols, Y., Pritzel, A., Eslami, S.~A., and Viola, F.
\newblock Generative temporal models with spatial memory for partially observed
  environments.
\newblock In \emph{International Conference on Machine Learning}, pp.\
  1544--1553, 2018.

\bibitem[Garnelo et~al.(2018{\natexlab{a}})Garnelo, Rosenbaum, Maddison,
  Ramalho, Saxton, Shanahan, Teh, Rezende, and Eslami]{garnelo2018conditional}
Garnelo, M., Rosenbaum, D., Maddison, C.~J., Ramalho, T., Saxton, D., Shanahan,
  M., Teh, Y.~W., Rezende, D.~J., and Eslami, S.
\newblock Conditional neural processes.
\newblock \emph{arXiv preprint arXiv:1807.01613}, 2018{\natexlab{a}}.

\bibitem[Garnelo et~al.(2018{\natexlab{b}})Garnelo, Schwarz, Rosenbaum, Viola,
  Rezende, Eslami, and Teh]{garnelo2018neural}
Garnelo, M., Schwarz, J., Rosenbaum, D., Viola, F., Rezende, D.~J., Eslami, S.,
  and Teh, Y.~W.
\newblock Neural processes.
\newblock \emph{arXiv preprint arXiv:1807.01622}, 2018{\natexlab{b}}.

\bibitem[Gemici et~al.(2017)Gemici, Hung, Santoro, Wayne, Mohamed, Rezende,
  Amos, and Lillicrap]{gemici2017generative}
Gemici, M., Hung, C.-C., Santoro, A., Wayne, G., Mohamed, S., Rezende, D.~J.,
  Amos, D., and Lillicrap, T.
\newblock Generative temporal models with memory.
\newblock \emph{arXiv preprint arXiv:1702.04649}, 2017.

\bibitem[Goyal et~al.(2017)Goyal, Sordoni, C{\^o}t{\'e}, Ke, and
  Bengio]{goyal2017z}
Goyal, A. G. A.~P., Sordoni, A., C{\^o}t{\'e}, M.-A., Ke, N.~R., and Bengio, Y.
\newblock Z-forcing: Training stochastic recurrent networks.
\newblock In \emph{Advances in neural information processing systems}, pp.\
  6713--6723, 2017.

\bibitem[Gregor et~al.(2016)Gregor, Besse, Rezende, Danihelka, and
  Wierstra]{gregor2016towards}
Gregor, K., Besse, F., Rezende, D.~J., Danihelka, I., and Wierstra, D.
\newblock Towards conceptual compression.
\newblock In \emph{Advances In Neural Information Processing Systems}, pp.\
  3549--3557, 2016.

\bibitem[Hafner et~al.(2018)Hafner, Lillicrap, Fischer, Villegas, Ha, Lee, and
  Davidson]{hafner2018learning}
Hafner, D., Lillicrap, T., Fischer, I., Villegas, R., Ha, D., Lee, H., and
  Davidson, J.
\newblock Learning latent dynamics for planning from pixels.
\newblock \emph{arXiv preprint arXiv:1811.04551}, 2018.

\bibitem[Karl et~al.(2016)Karl, Soelch, Bayer, and van~der Smagt]{karl2016deep}
Karl, M., Soelch, M., Bayer, J., and van~der Smagt, P.
\newblock Deep variational bayes filters: Unsupervised learning of state space
  models from raw data.
\newblock \emph{arXiv preprint arXiv:1605.06432}, 2016.

\bibitem[Kim et~al.(2019)Kim, Mnih, Schwarz, Garnelo, Eslami, Rosenbaum,
  Vinyals, and Teh]{kim2019attentive}
Kim, H., Mnih, A., Schwarz, J., Garnelo, M., Eslami, A., Rosenbaum, D.,
  Vinyals, O., and Teh, Y.~W.
\newblock Attentive neural processes.
\newblock \emph{arXiv preprint arXiv:1901.05761}, 2019.

\bibitem[Kingma \& Welling(2013)Kingma and Welling]{kingma2013auto}
Kingma, D.~P. and Welling, M.
\newblock Auto-encoding variational bayes.
\newblock \emph{arXiv preprint arXiv:1312.6114}, 2013.

\bibitem[Krishnan et~al.(2017)Krishnan, Shalit, and
  Sontag]{krishnan2017structured}
Krishnan, R.~G., Shalit, U., and Sontag, D.
\newblock Structured inference networks for nonlinear state space models.
\newblock In \emph{Thirty-First AAAI Conference on Artificial Intelligence},
  2017.

\bibitem[Kumar et~al.(2018)Kumar, Eslami, Rezende, Garnelo, Viola, Lockhart,
  and Shanahan]{kumar2018consistent}
Kumar, A., Eslami, S., Rezende, D.~J., Garnelo, M., Viola, F., Lockhart, E.,
  and Shanahan, M.
\newblock Consistent generative query networks.
\newblock \emph{arXiv preprint arXiv:1807.02033}, 2018.

\bibitem[Mordatch et~al.()Mordatch, Lowrey, and Todorov]{mordatch2015ensemble}
Mordatch, I., Lowrey, K., and Todorov, E.
\newblock Ensemble-cio: Full-body dynamic motion planning that transfers to
  physical humanoids.
\newblock In \emph{2015 IEEE/RSJ International Conference on Intelligent Robots
  and Systems (IROS)}, pp.\  5307--5314. IEEE.

\bibitem[Nair \& Hinton(2010)Nair and Hinton]{nair2010rectified}
Nair, V. and Hinton, G.~E.
\newblock Rectified linear units improve restricted boltzmann machines.
\newblock In \emph{Proceedings of the 27th international conference on machine
  learning (ICML-10)}, pp.\  807--814, 2010.

\bibitem[Rosenbaum et~al.(2018)Rosenbaum, Besse, Viola, Rezende, and
  Eslami]{rosenbaum2018learning}
Rosenbaum, D., Besse, F., Viola, F., Rezende, D.~J., and Eslami, S.
\newblock Learning models for visual 3d localization with implicit mapping.
\newblock \emph{arXiv preprint arXiv:1807.03149}, 2018.

\bibitem[Srivastava et~al.(2015)Srivastava, Mansimov, and
  Salakhudinov]{srivastava2015unsupervised}
Srivastava, N., Mansimov, E., and Salakhudinov, R.
\newblock Unsupervised learning of video representations using lstms.
\newblock In \emph{International conference on machine learning}, pp.\
  843--852, 2015.

\bibitem[Xingjian et~al.(2015)Xingjian, Chen, Wang, Yeung, Wong, and
  Woo]{xingjian2015convolutional}
Xingjian, S., Chen, Z., Wang, H., Yeung, D.-Y., Wong, W.-K., and Woo, W.-c.
\newblock Convolutional lstm network: A machine learning approach for
  precipitation nowcasting.
\newblock In \emph{Advances in neural information processing systems}, pp.\
  802--810, 2015.

\bibitem[Zheng et~al.(2017)Zheng, Zaheer, Ahmed, Wang, Xing, and
  Smola]{zheng2017state}
Zheng, X., Zaheer, M., Ahmed, A., Wang, Y., Xing, E.~P., and Smola, A.~J.
\newblock State space lstm models with particle mcmc inference.
\newblock \emph{arXiv preprint arXiv:1711.11179}, 2017.

\end{thebibliography}

\newpage
\begin{appendices}

\newpage
\section{Additional Demonstrations of SNP}
In this section, we show additional qualitative demonstrations of SNP and comparisons against NP, GQN and the ground truth.
\subsection{Uncertainty modeling and meta-transfer learning in SNP}
\label{ax:unc_demo}

\begin{figure}[h!]
    \centering
    \begin{subfigure}{1.0\textwidth}
        \centering
        \includegraphics[width=1.0\linewidth]{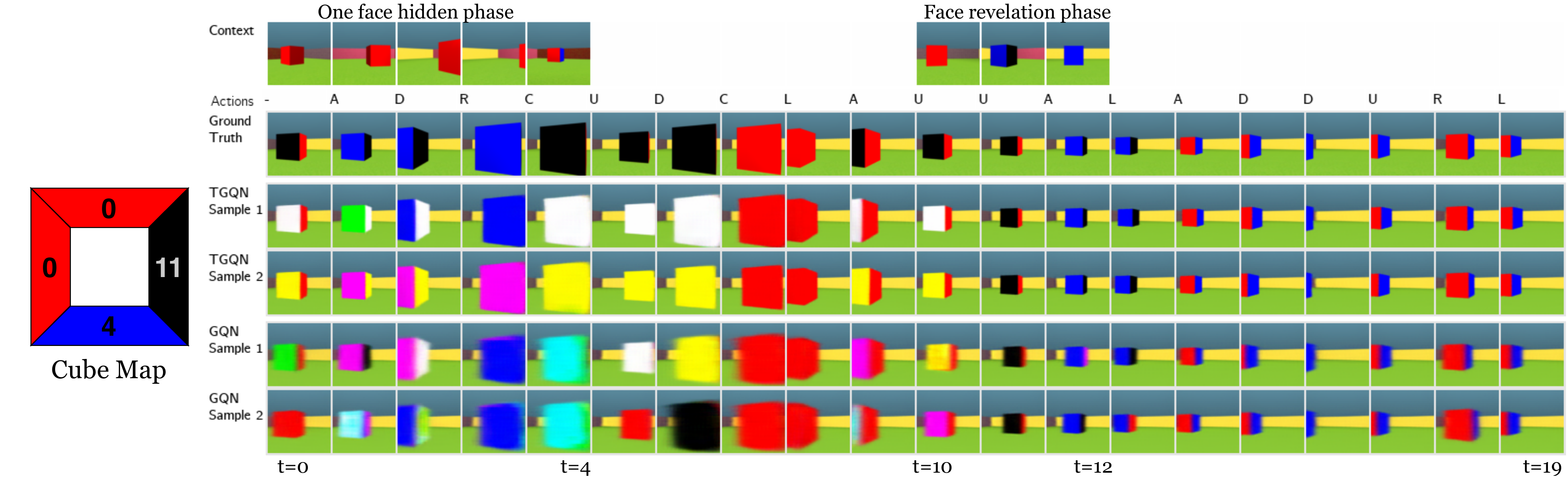}
        \caption{}
    \end{subfigure}
    \begin{subfigure}{1.0\textwidth}
        \centering
        \includegraphics[width=1.0\linewidth]{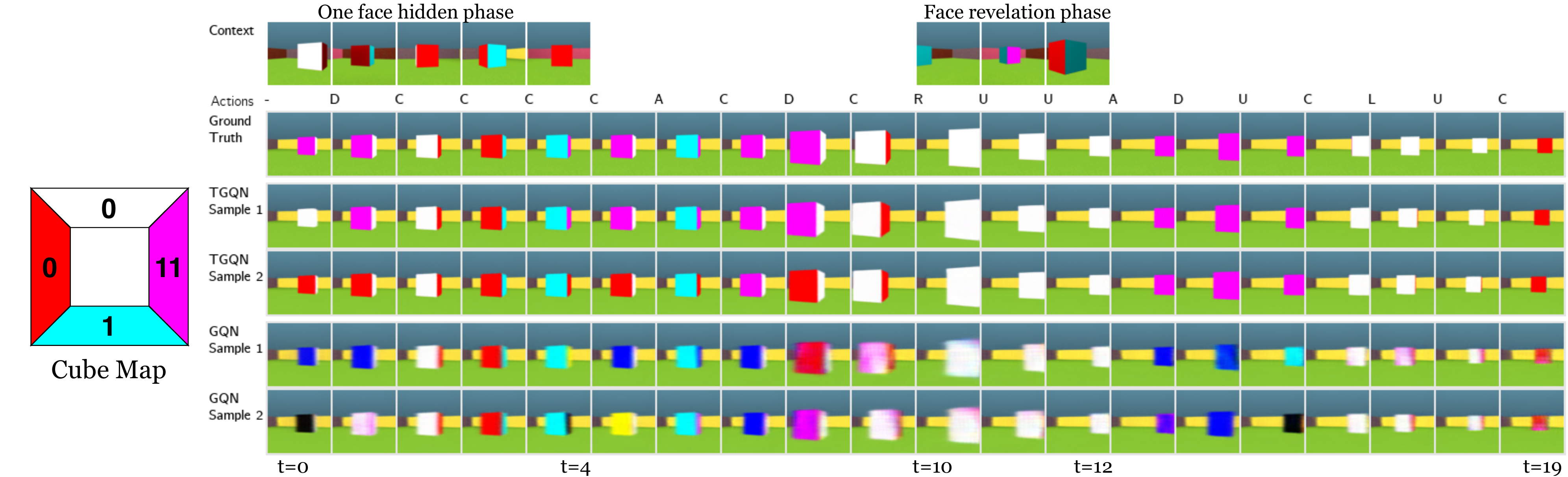}
        \caption{}
    \end{subfigure}
    \begin{subfigure}{1.0\textwidth}
        \centering
        \includegraphics[width=1.0\linewidth]{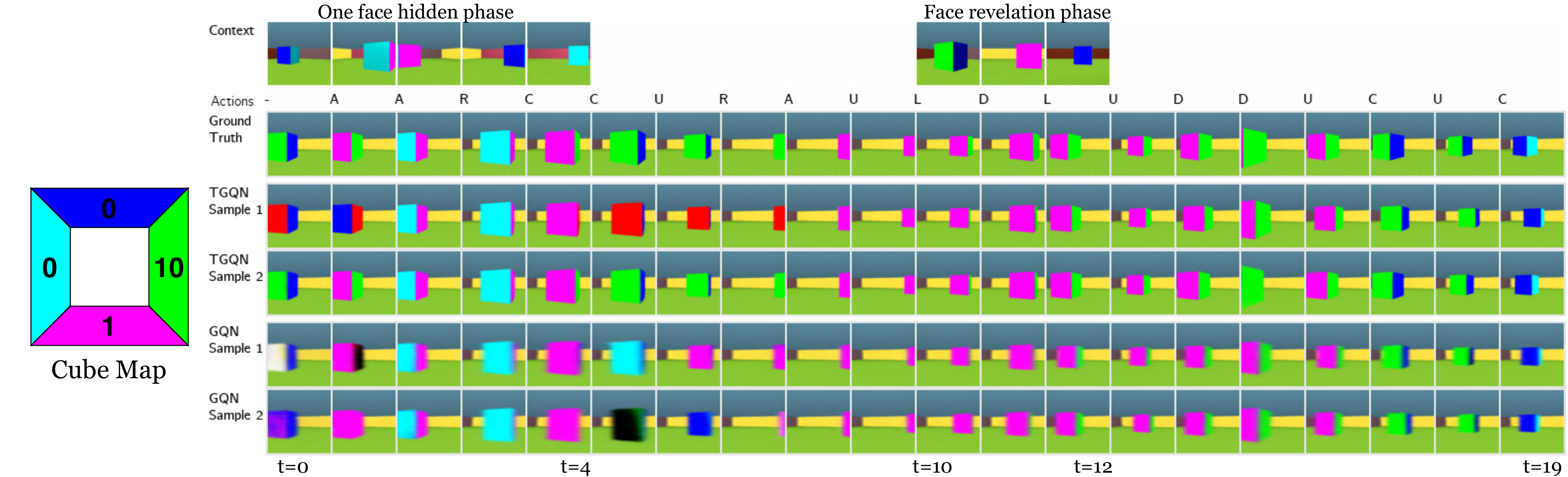}
        \caption{}
    \end{subfigure}
    \caption{Goal of this demonstration is to show uncertainty modeling and meta-transfer learning in TGQN and GQN in the jittery color-cube environment. We provide contexts in two phases. In the early phase ($t=0$ to $4$), we show one observation per time-step while avoiding revealing a particular face. In the late phase ($t=10$ to $12$), we reveal that face. On the left, we show true face colors of the cube with the numbers showing the time-step at which it is first revealed. We generate two samples of rollouts of TGQN and GQN each. We make the following observations. \emph{i)} In time-steps 5 through 9, we observe that TGQN can model uncertainty when the face colors are unseen and samples a color from the true palette.
    \emph{ii)} In time-steps 10 through 12, we note that the face revelation updates the color of the previously unseen face. Since the cube dynamics are jittery, we also note that this context re-synchronises the cube position. Furthermore, we observe that TGQN transfers its knowledge of previously seen faces and combines it with the newly revealed face, thus performing meta-transfer learning. This new knowledge is maintained in the predictions made henceforth.
    \emph{iii)} Overall, GQN produces blurred generations with inconsistency in the colors of the unseen faces.}
\end{figure}

\begin{figure}[h!]
    \centering
    \begin{subfigure}{1.0\textwidth}
        \centering
        \includegraphics[width=1.0\linewidth]{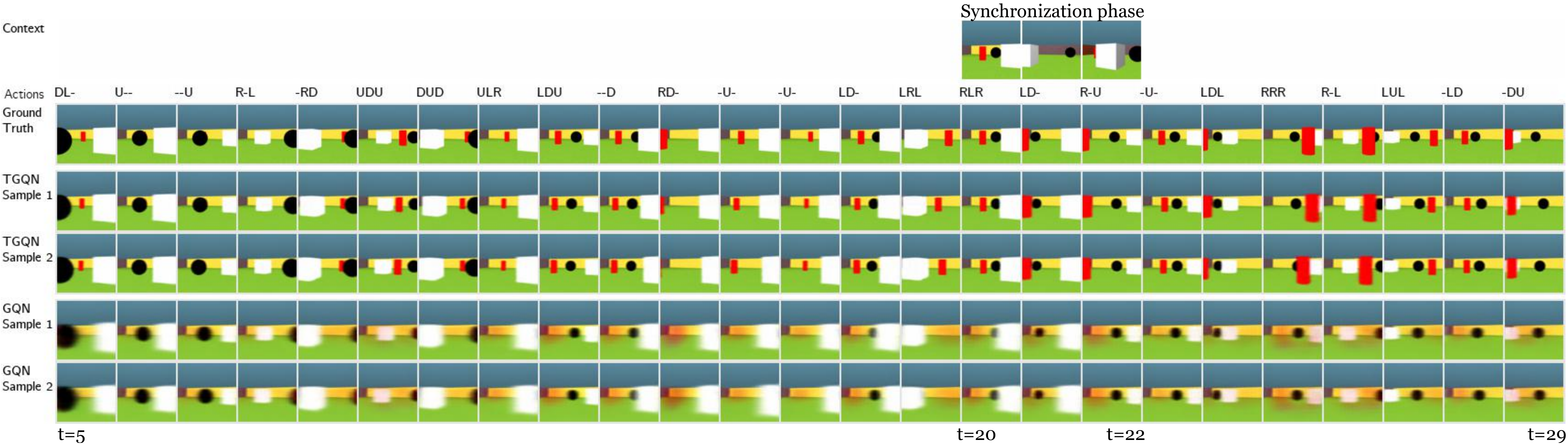}
        \caption{}
    \end{subfigure}
    \begin{subfigure}{1.0\textwidth}
        \centering
        \includegraphics[width=1.0\linewidth]{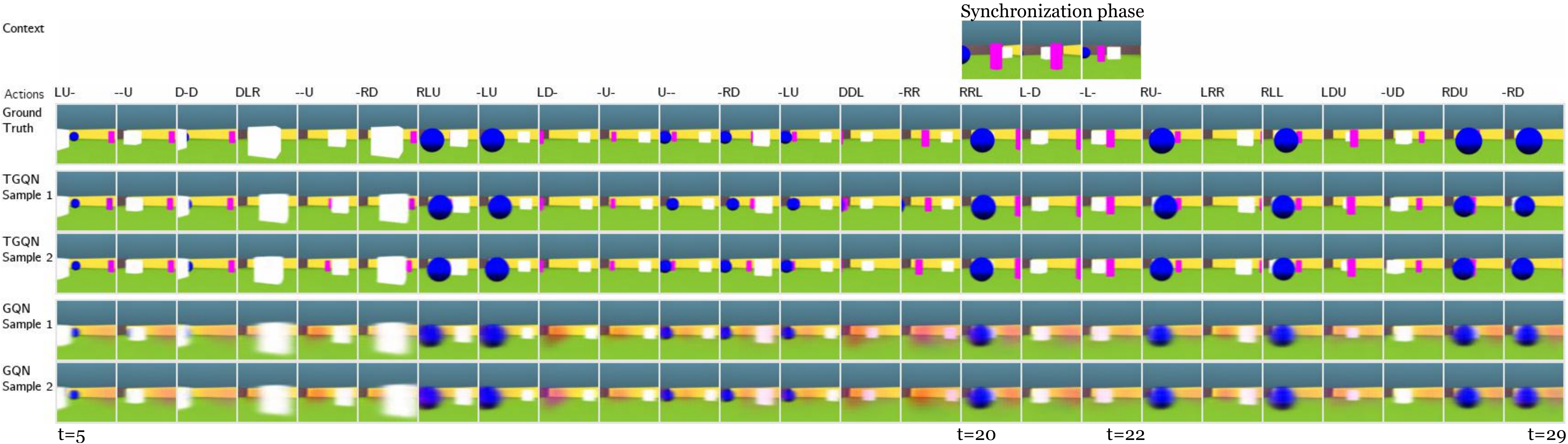}
        \caption{}
    \end{subfigure}
    \begin{subfigure}{1.0\textwidth}
        \centering
        \includegraphics[width=1.0\linewidth]{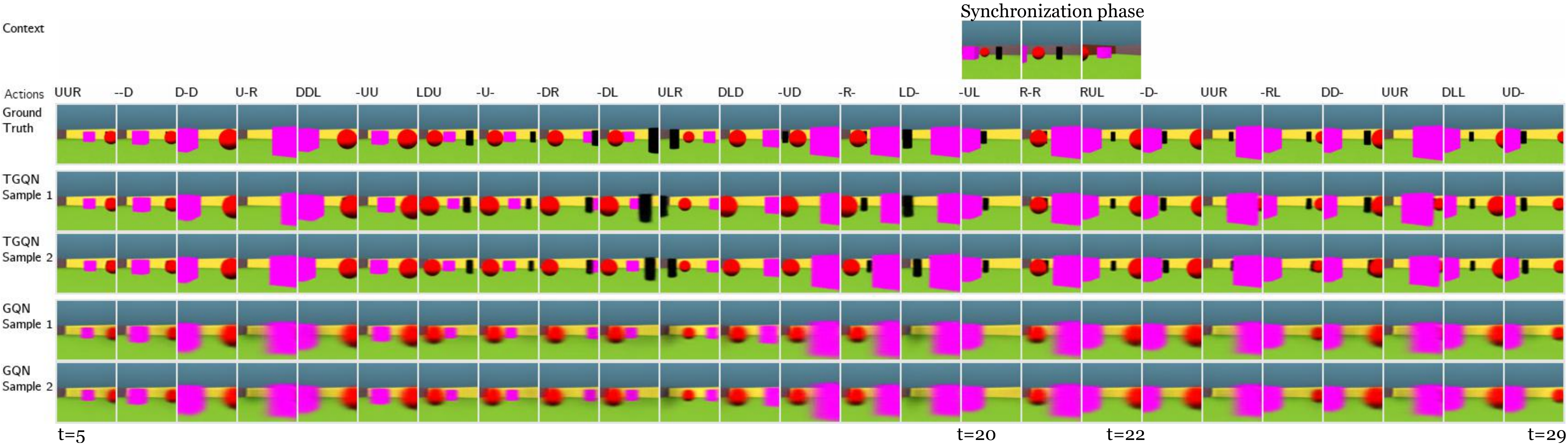}
        \caption{}
    \end{subfigure}
    \caption{Goal of this demonstration is to show tracking ability of TGQN versus GQN in the jittery multi-object environment. We provide contexts in two phases. In the early phase ($t=0$ to 4), we show one context observation at each time-step. Then we let the model predict the next 15 time-steps so that the predictions diverge from the true scene due to the jitter in the dataset. In the synchronisation phase ($t=20$ to 22), we show one observation per time-step. In this demonstration, we show two samples of the rollout from TGQN and GQN each from time-steps 5 through 29. We make the following observations. \emph{i)} In time-steps 5 through 19, TGQN shows that it appropriately models the transition stochasticity as different samples produce different object positions at $t=19$. \emph{ii)} At $t=20$, we see that the context re-synchronises the object positions with the true positions. \emph{iii)} Overall, GQN produces blurred generations and is not able to model the cylinders.}

\end{figure}

\begin{figure}[h!]
    \centering
    \begin{subfigure}{1.0\textwidth}
        \centering
        \includegraphics[width=1.0\linewidth]{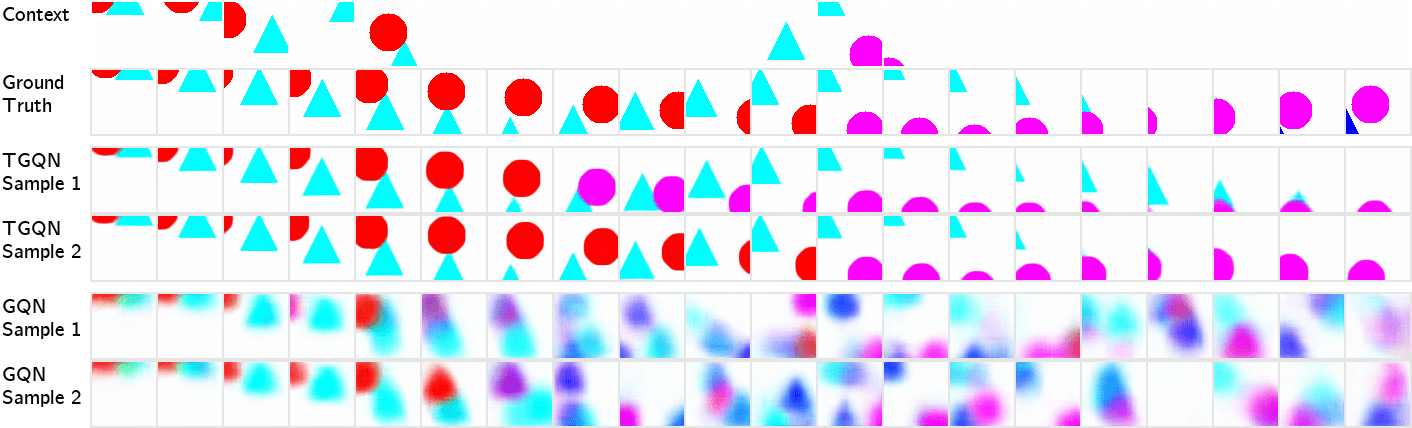}
        \caption{}
    \end{subfigure}
    \begin{subfigure}{1.0\textwidth}
        \centering
        \includegraphics[width=1.0\linewidth]{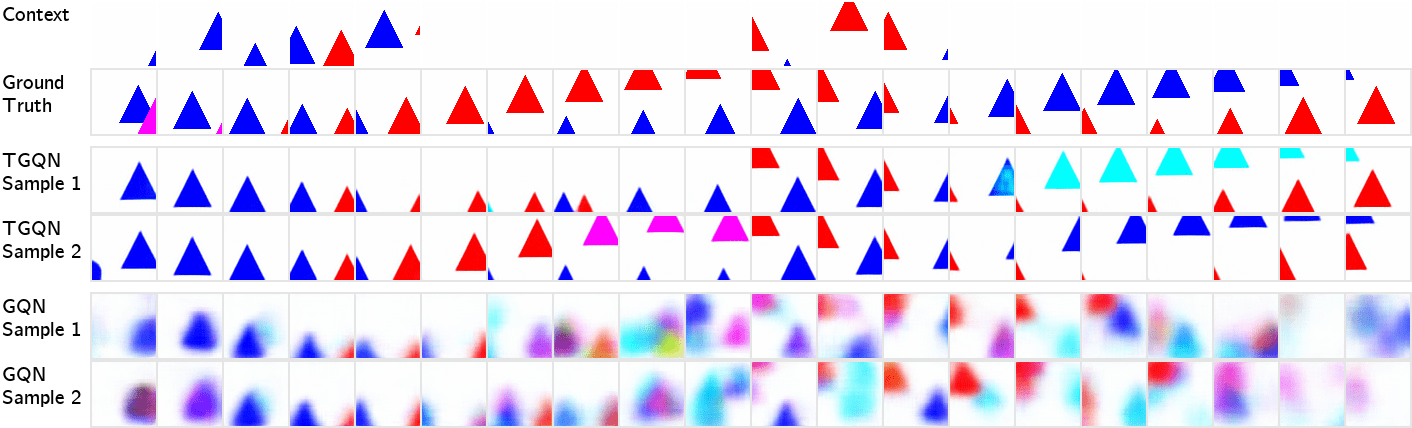}
        \caption{}
    \end{subfigure}
    \begin{subfigure}{1.0\textwidth}
        \centering
        \includegraphics[width=1.0\linewidth]{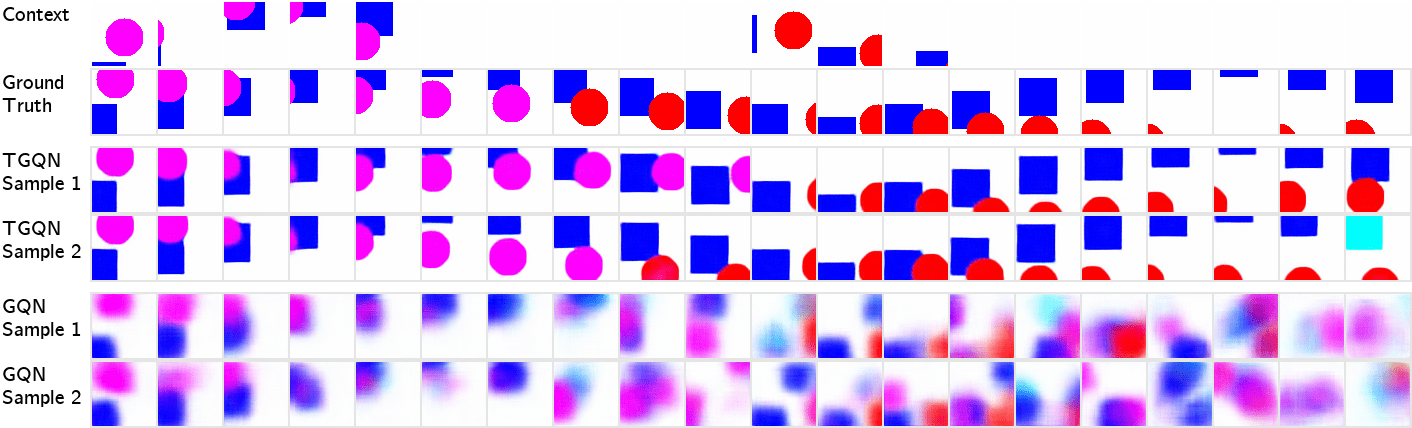}
        \caption{}
    \end{subfigure}
    \caption{Goal of this demonstration is to show tracking ability of TGQN versus GQN in the 2D color-shapes dataset. We provide contexts in two phases. In the early phase ($t=0$ to 4), we show one context observation at each time-step. Then we let the model predict the next 5 time-steps so that the predictions diverge from the true scene due to the random color-change in the dataset. In the synchronisation phase ($t=10$ to 12), we show one observation per time-step. In this demonstration, we show two samples of the rollout from TGQN and GQN each from time-steps 0 through 19. We make the following observations. \emph{i)} In time-steps 5 through 9, TGQN shows that it appropriately models the transition stochasticity as different samples produce different object positions and color-changes. \emph{ii)} At $t=10$, we see that the context re-synchronises the object positions and colors with the true ones. \emph{iii)} Overall, GQN produces blurred generations.}

\end{figure}

\begin{figure}[h!]
    \centering
    \begin{subfigure}{0.49\textwidth}
        \centering
        \includegraphics[width=1.0\linewidth]{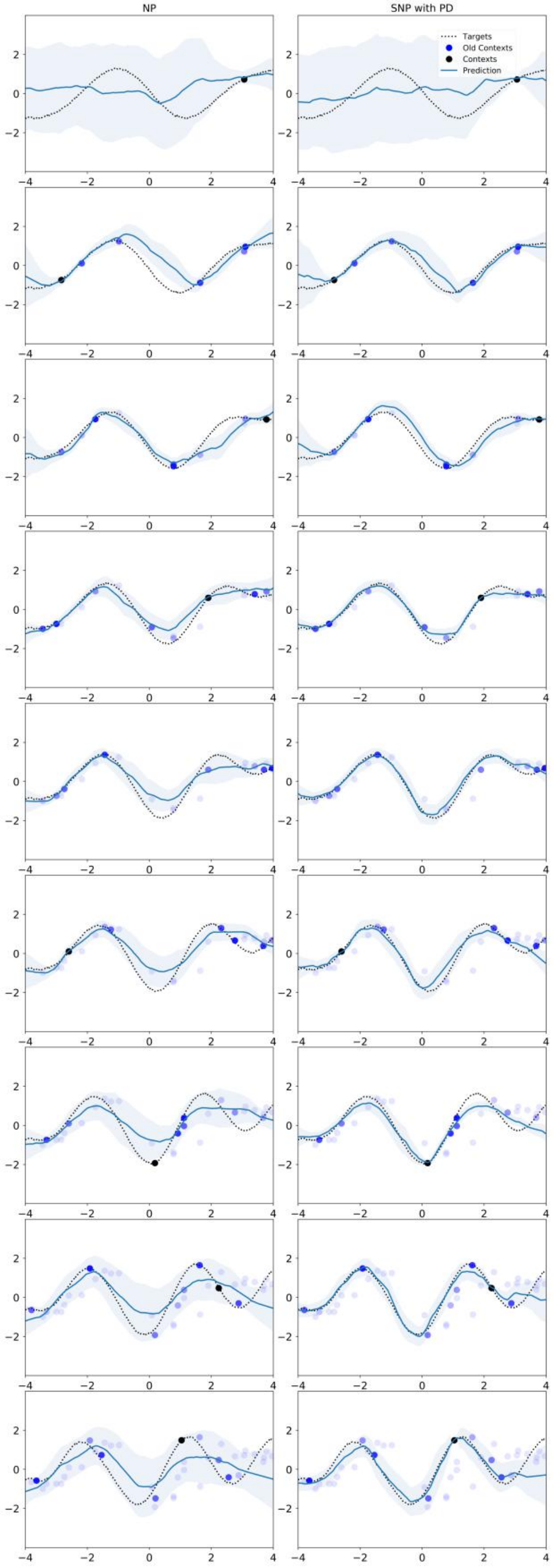}
        \caption{Episode 1}
    \end{subfigure}
    \begin{subfigure}{0.49\textwidth}
        \centering
        \includegraphics[width=1.0\linewidth]{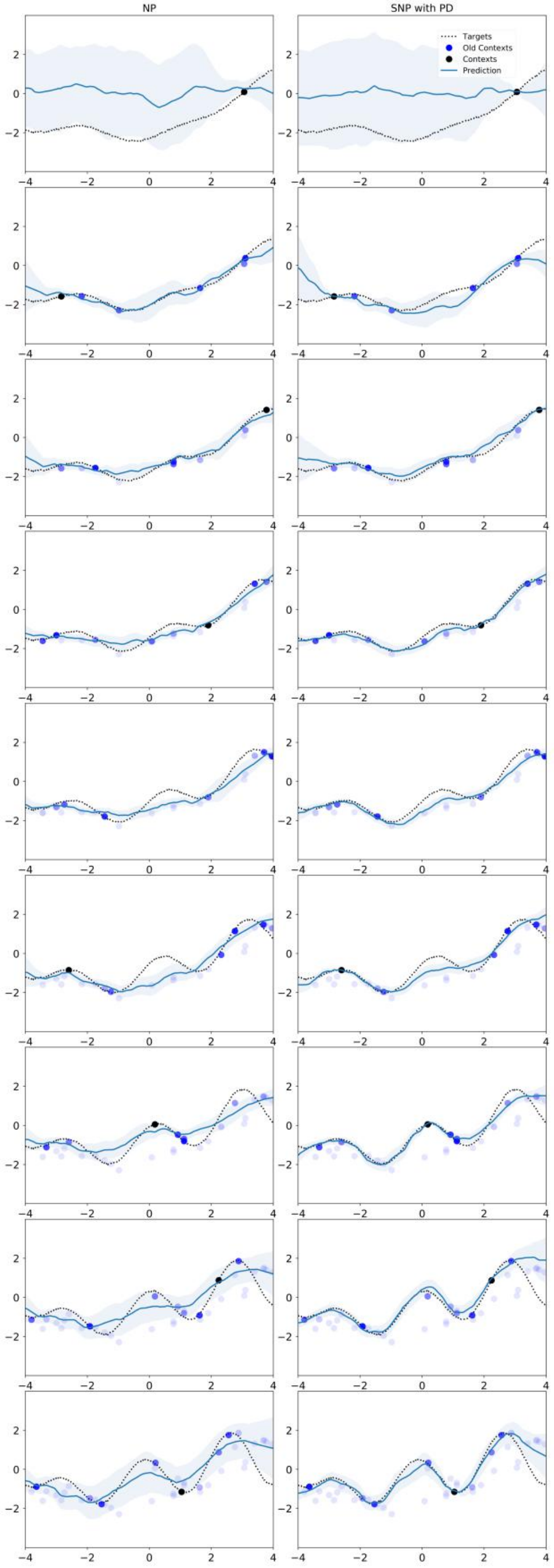}
        \caption{Episode 2}
    \end{subfigure}
    \caption{1D regression qualitative samples for task (c). Each row corresponds to one time-step. Due to space limitations, every $5^{\text{th}}$ time-step is shown here instead of every time-step up to 45.}
    \label{fig:reg_gen_vz}
\end{figure}

\clearpage
\FloatBarrier
\subsection{Prediction in SNP}
In this section, we demonstrate the predictions using SNP. 
\label{sec:comp_examples}
\begin{figure}[h!]
    \centering
    \begin{subfigure}{0.94\textwidth}
        \centering
        \includegraphics[width=1.0\linewidth]{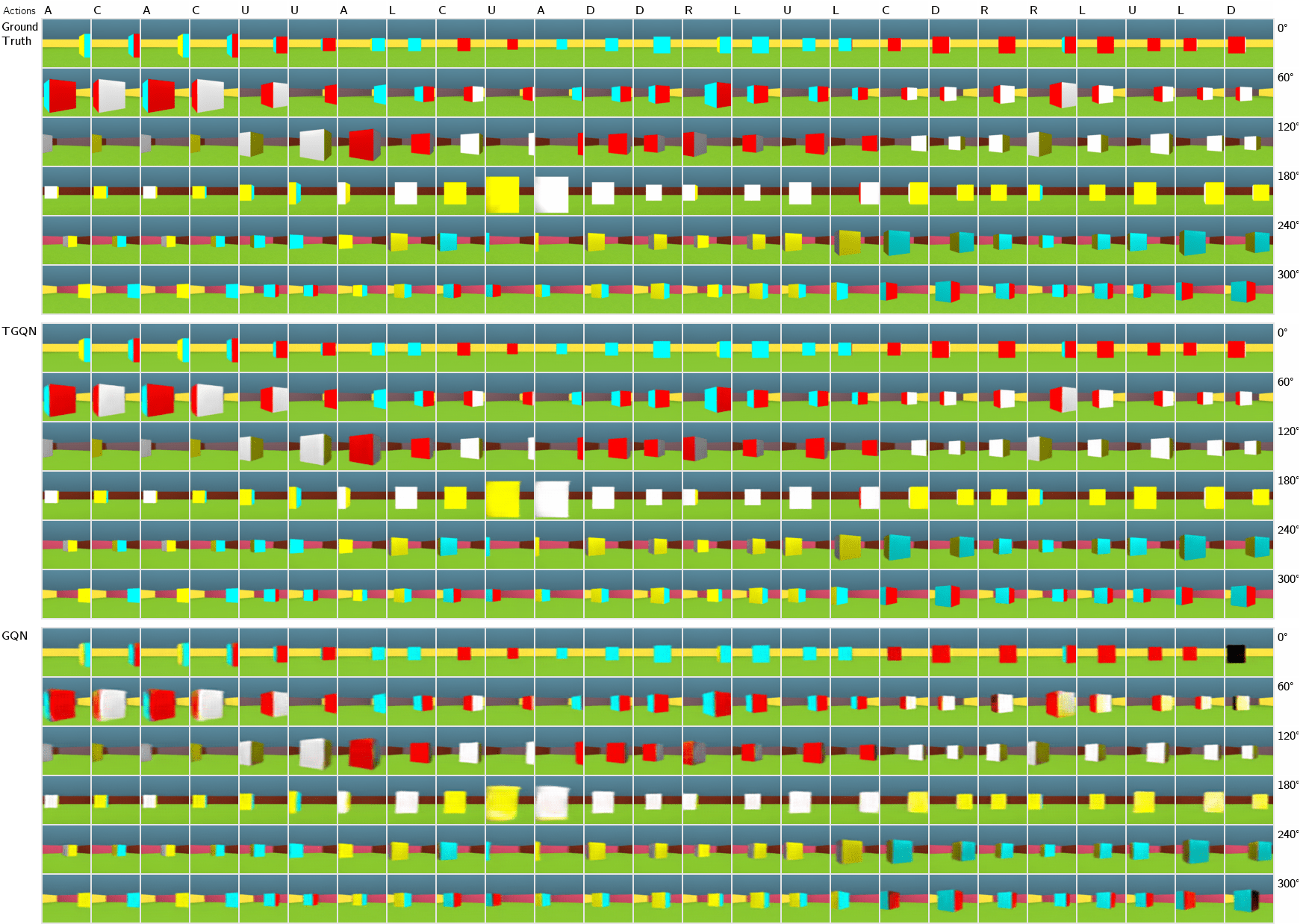}
    \end{subfigure}
    \begin{subfigure}{0.94\textwidth}
        \centering
        \includegraphics[width=1.0\linewidth]{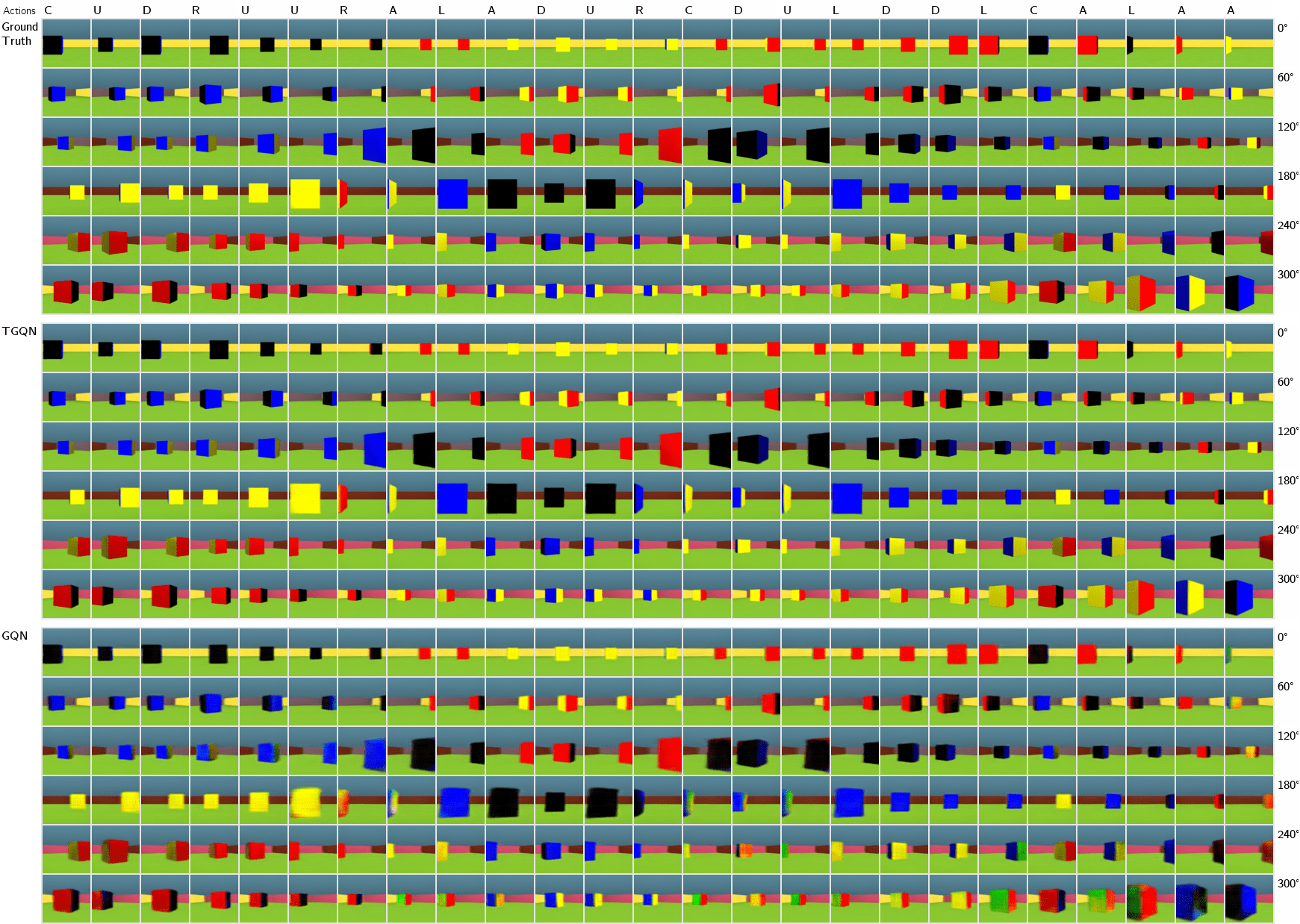}
    \end{subfigure}
    \caption{The goal of this demonstration is to show predictions from $t=5$ through 29 using the context shown only in the early time-steps $t<5$. Each row shows views from cameras positioned at angles labelled on the right. We compare TGQN with GQN and the ground truth. We observe that TGQN makes clear predictions even beyond the training sequence length $T=10$. In contrast, GQN's generations are blurred with susceptibility to forgetting face colors.}
\end{figure}

\begin{figure}[h!]
    \centering
    \begin{subfigure}{1.0\textwidth}
        \centering
        \includegraphics[width=1.0\linewidth]{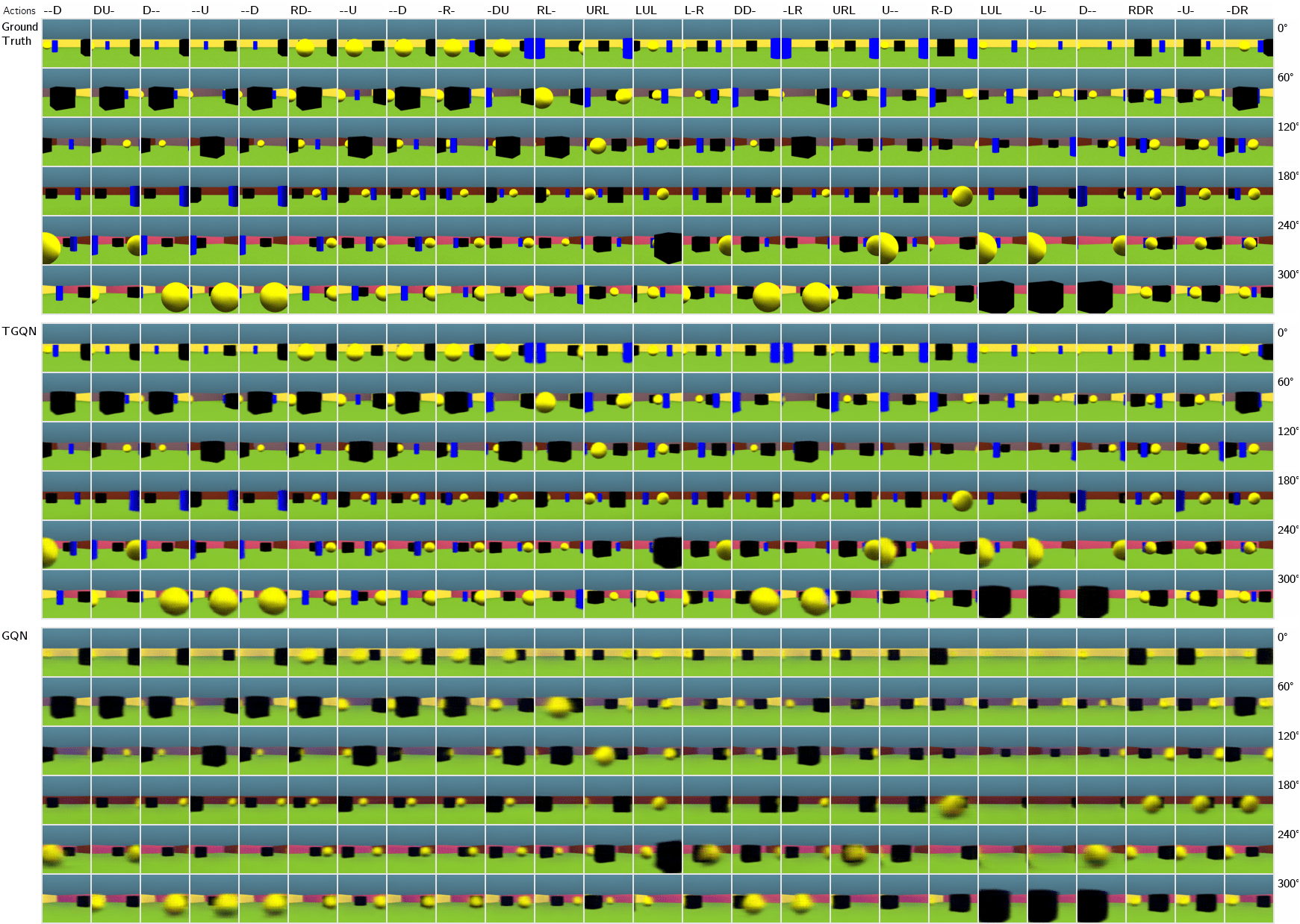}
    \end{subfigure}
    \begin{subfigure}{1.0\textwidth}
        \centering
        \includegraphics[width=1.0\linewidth]{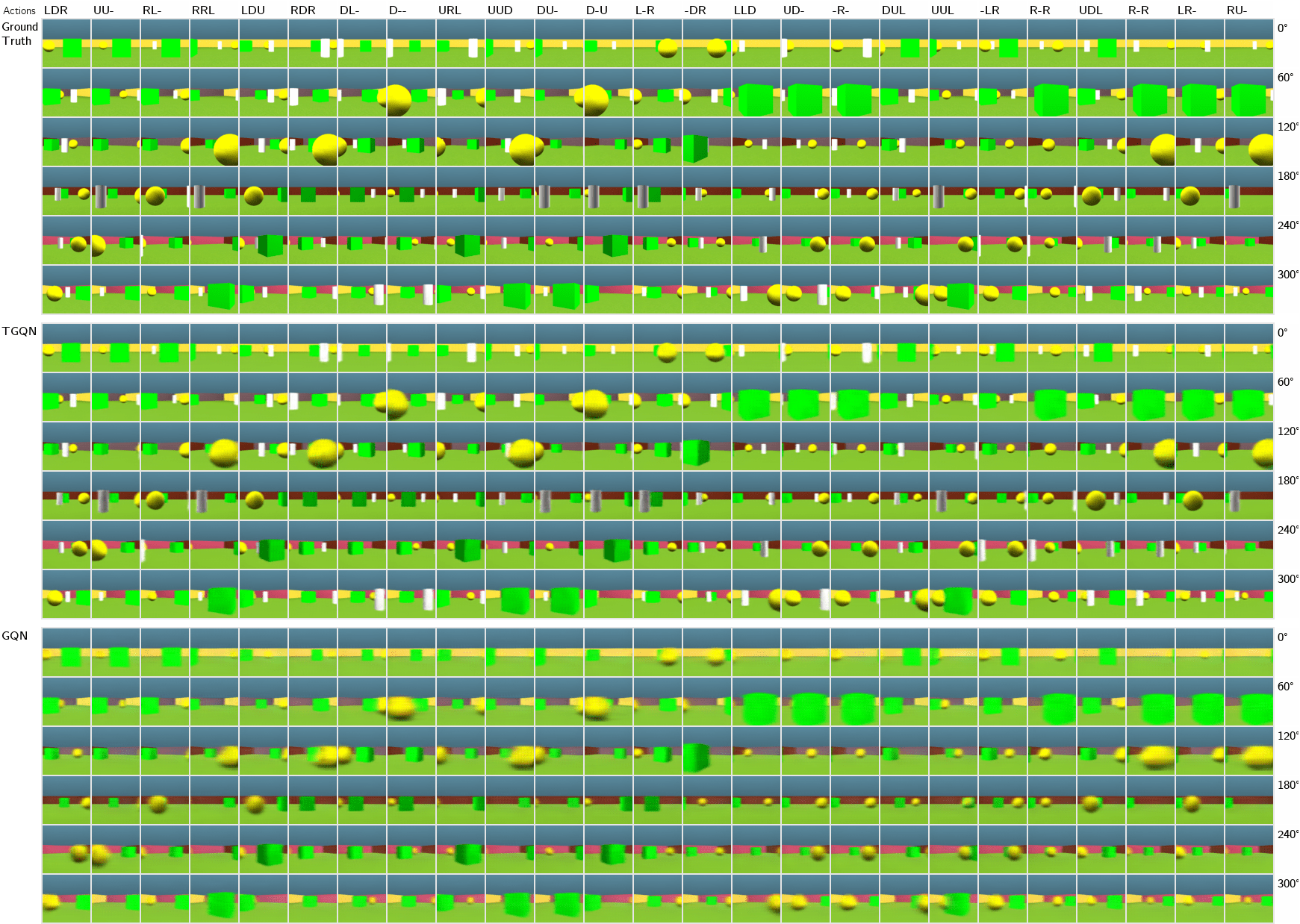}
    \end{subfigure}
    \caption{The goal of this demonstration is to show predictions from $t=5$ through 29 using the context shown only in the early time-steps $t<5$. Each row shows views from cameras positioned at angles labelled on the right. We compare TGQN with GQN and the ground truth. We observe that TGQN makes clear predictions even beyond the training sequence length $T=10$. In contrast, GQN's generations are blurred and it also cannot model the finer details like the cylinder.}
\end{figure}

\begin{figure}[h!]
    \centering
    \begin{subfigure}{1.0\textwidth}
        \centering
        \includegraphics[width=1.0\linewidth]{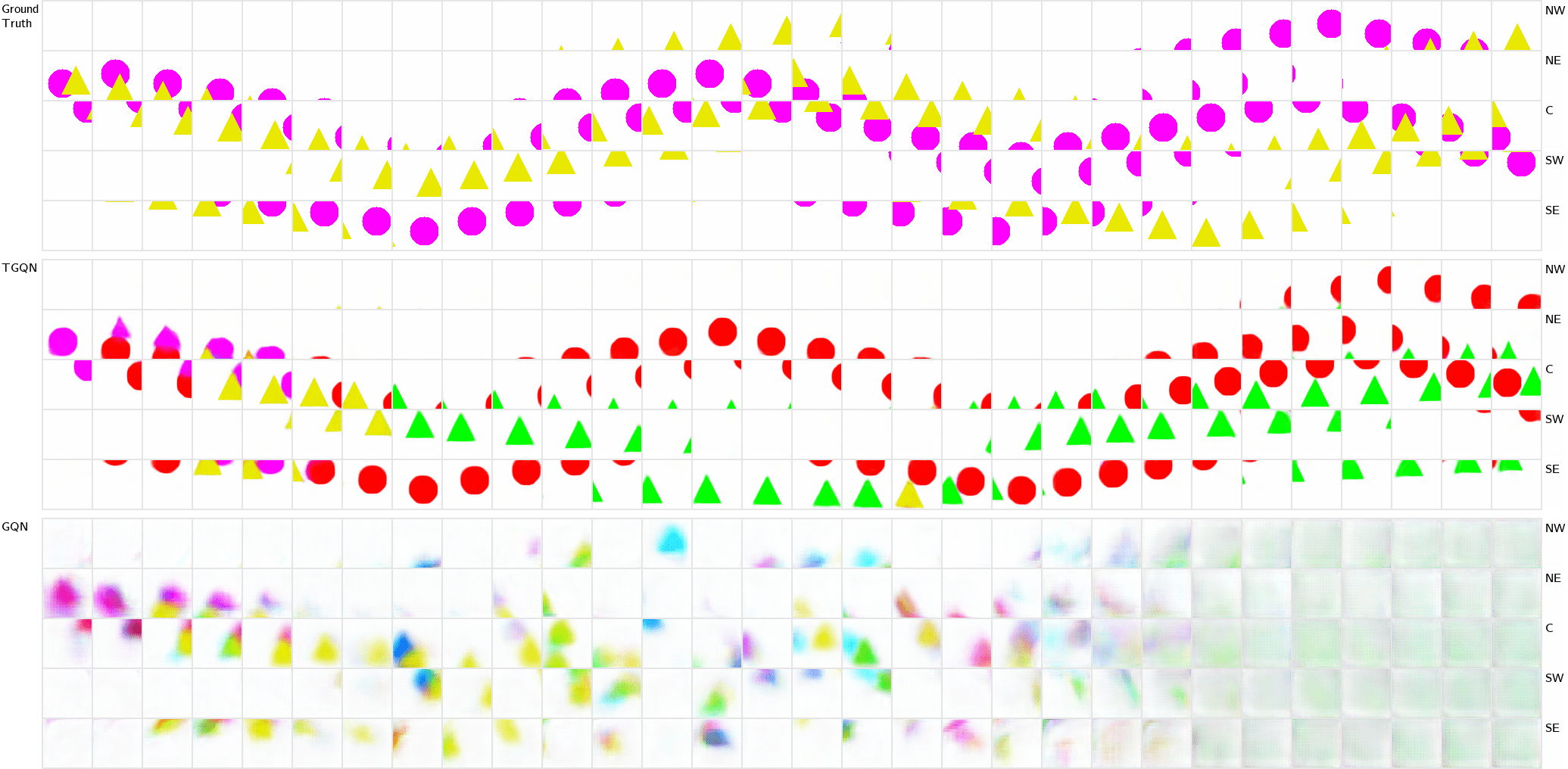}
    \end{subfigure}
    \begin{subfigure}{1.0\textwidth}
        \centering
        \includegraphics[width=1.0\linewidth]{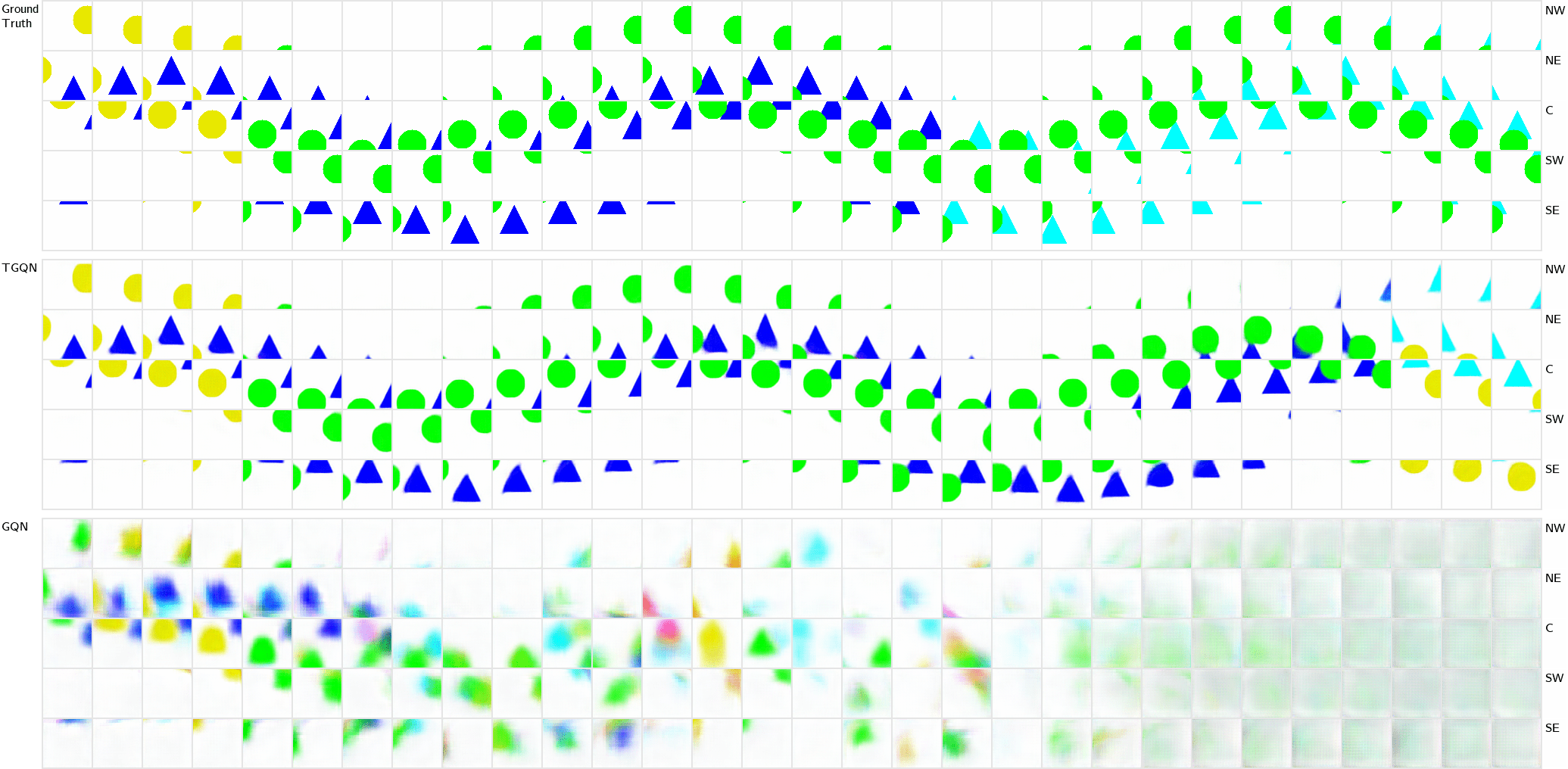}
    \end{subfigure}
    \caption{The goal of this demonstration is to show predictions from $t=0$ through 29 using the context shown only in the early time-steps $t<5$. We compare TGQN with GQN and the ground truth. The labels on the right show where the queried patch is located e.g.,~\textit{SW} indicates that the query patch is located at south-west corner of the canvas. We observe in time-steps $t\geq5$, when the context is removed, that TGQN keeps making plausible predictions in accordance with the stochastic color change rules and motion dynamics. This continues even beyond the training sequence length $T=20$. In contrast, GQN's predictions are blurred and cannot generalize beyond $T=20$.}
\end{figure}

\FloatBarrier
\clearpage
\section{ELBO Derivations}
In this section, we derive the ELBO expressions that were introduced in the main text of the paper.
\subsection{SNP ELBO}
\label{sec:snp_elbo_deriv}
In this sub-section we derive the ELBO mentioned in~\eqref{eq:snp_elbo}. We start with the objective of maximizing the log-likelihood of the targets given the queries and the contexts.
\begin{flalign*}
\begin{aligned}
&\log P(Y|X,C)\\ 
  &= \log \eE_{Q_\phi(Z|\cV)} \frac{P(Y,Z|X,C)}{Q_\phi(Z|\cV)}\\ 
  &= \log \eE_{Q_\phi(Z|\cV)} \frac{\pd{t}{T}P_\ta(Y_t|X_t,z_t)P_\ta(z_t|z_\lt,C_t)}{\pd{t}{T}Q_\phi(z_t|z_{<t},\cV)}  \\
&\geq \eE_{Q_\phi(Z|\cV)}  \left[ \log \frac{\pd{t}{T}P_\ta(Y_t|X_t,z_t)P_\ta(z_t|z_\lt,C_t)}{\pd{t}{T}Q_\phi(z_t|z_{<t},\cV)}   \right]\\
&= \eE_{Q_\phi(Z|\cV)}  \sm{t}{T}\left[ \log \frac{P_\ta(Y_t|X_t,z_t)P_\ta(z_t|z_\lt,C_t)}{Q_\phi(z_t|z_{<t},\cV)}   \right]\\
&= \eE_{Q_\phi(Z|\cV)}  \sm{t}{T}\left[\log P_\ta(Y_t|X_t,z_t) + \log \frac{P_\ta(z_t|z_\lt,C_t)}{Q_\phi(z_t|z_{<t},\cV)}  \right]\\
 &= \sm{t}{T}\eE_{Q_\phi(Z|\cV)}\left[\log P_\ta(Y_t|X_t,z_t) + \log \frac{P_\ta(z_t|z_\lt,C_t)}{Q_\phi(z_t|z_{<t},\cV)}   \right]\\
 &= \sm{t}{T}\eE_{Q_\phi(Z|\cV)}\left[\log P_\ta(Y_t|X_t,z_t) - \log \frac{Q_\phi(z_t|z_{<t},\cV) }{P_\ta(z_t|z_\lt,C_t)} \right]\\
 &=\sm{t}{T}\eE_{Q_\phi(z_t|\cV)}\left[\log P_\ta(Y_t|X_t,z_t)\right] - \eE_{Q_\phi(z_{\leq t}|\cV)}\log \frac{Q_\phi(z_t|z_{<t},\cV) }{P_\ta(z_t|z_\lt,C_t)} \\
 &=\sm{t}{T}\eE_{Q_\phi(z_t|\cV)}\left[\log P_\ta(Y_t|X_t,z_t)\right] - \eE_{Q_\phi(z_\lt|\cV)}\left[\KL(Q_\phi(z_t|z_{<t},\cV) \parallel P_\ta(z_t|z_\lt,C_t)) \right]
\end{aligned}
\end{flalign*}
which gives us the expression in~\eqref{eq:snp_elbo}.

\subsection{Posterior Dropout ELBO}
In this sub-section, we derive the ELBO with \emph{posterior dropout}~\eqref{eq:aux_elbo}. As mentioned in Section~\ref{sec:posterior_dropout}, we choose a subset of time-steps $\cT$ so that we use the prior distribution to sample the $z_t$ and posterior for the time-steps in $\tilde{\cT}$. We start with the objective of maximizing the likelihood of the target images belonging to the time-steps in $\tilde{\cT}$ and then proceed with  the derivation as shown below.
\label{sec:aux_elbo_deriv}
\begin{flalign*}
    \eE_{\tilde{\cT}}&\log{P_{\ta}(Y_{\bar{\cT}}|X,C)} \\
    &= \eE_{\tilde{\cT}}\log\int  \prod_{t\in\bar{\cT}}P_\ta(y_t|x_t,z_{t}) \pd{t}{T}P_\ta(z_t|z_{<t},C_t) dZ\\
    &= \eE_{\tilde{\cT}}\log\E_{Z\sim \tilde{Q}}\left[ \frac{\prod_{t\in\bar{\cT}}P_\ta(y_t|x_t,z_{t}) \pd{t}{T}P_\ta(z_t|z_{<t},C_t)}{\prod_{t \in \cT}P_\ta(z_t|z_{<t},C_t)\prod_{t \in \bar{\cT}} Q_\phi(z_t|z_{<t},C,D)} \right]\\
    &\geq \eE_{\tilde{\cT}}\E_{Z\sim \tilde{Q}}\sum_{t \in \bar{\cT}}\left[ \log P_\ta(y_t|x_t,z_{t}) -  \KL\left( Q_\phi(z_t|z_{<t},C,D) \parallel P_\ta(z_t|z_{<t},C_t) \right)\right] = \mathcal{L}_{\text{PD}} 
\end{flalign*}
which gives us the required expression in~\eqref{eq:aux_elbo}.

\section{Neural Networks}

\subsection{Sequential Neural Processes and the baseline Neural Processes for 1D regression} \label{1d_model}
\label{ax:1d-reg-snp-np-nn}
SNP and the NP baseline have two encoders: \textit{deterministic encoder} and \textit{latent encoder}. This model does not consume actions. The deterministic encoder consists of a 6-layer MLP with ReLU~\citep{nair2010rectified} activation. The latent encoder consists of a 3-layer MLP with ReLU followed a 2-layer MLP for computing sufficient statistics of the latent. This encoder acts as a prior when provided only with the context set, but also acts as the posterior when provided with the target set. We implement the state-space model using an LSTM with the default Tensorflow~\citep{abadi2016tensorflow} settings. 

Since NP is not a temporal architecture, normalized time $t' = 0.25 + 0.5\times(t/T)$ is appended to the original query $x$ to obtain $\tilde{x} = (x,t')$.

The dimension of the hidden units is 128. The learning rate and the batch size are 0.0001 and 16, respectively.

\subsection{Temporal Generative Query Networks}
\label{ax:tgqn-nn}
Here, we give the details of the implementation of the TGQN model geared towards generation of 3D scenes.
Our implementation is fully convolutional i.e., all the latent states and deterministic states are 3 dimensional tensors.

\paragraph{Generation} Below, we outline the implementation of the generative model.
\begin{flalign}
     h_0 &\leftarrow \text{learned parameter} && \text{(Initialize deterministic state)} \label{eq:h0}\\
     z_0 &\leftarrow \text{learned parameter}  && \text{(Initial latent)}\\
     C_t &\leftarrow \sum_{i=1}^{n_t} \repnet_\theta(x_i^t, y_i^t) && \text{(Compute scene representation from observed context)} \\
    a_t &\leftarrow \text{action embedding} && \text{(One-hot action embedding)}\\
    h_t &\leftarrow \rnn_\theta(h_{t-1}, z_{t-1}, a_{t-1}, C_t) && \text{(Deterministic state transition)} \label{eq:ht}\\
    z_t &\sim \draw_\theta(h_t, a_{t-1}, C_t) && \text{(Sample $z_t$ using DRAW)} \label{eq:zt_sample}\\
    y_i^t &\leftarrow \renderer_\gamma(x_i^t, z_t, h_t) && \text{(Render the image)}
\end{flalign}
More details about the implementation of $\draw_\theta$, $\repnet_\theta$ and the $\renderer_\gamma$ are provided in following sections.
\paragraph{Inference}
Next, we outline the inference procedure used for sampling all the latents $z_{1:T}$. 
\begin{align}
    D_t &\leftarrow \sum_{i=n_t + 1}^{m_t} \repnet_\theta(x_i^t, y_i^t) && \text{(Compute scene representation from target observations)} \\
    b_{t} &\leftarrow \rnn_\phi(b_{t+1}, C_t, D_t, a_t) && \text{(Encode all observations using a backward RNN)}\\
    z_t &\sim \draw_\phi(h_t, a_{t-1}, b_t) && \text{(Sample $z_t$ using DRAW.)} \label{eq:zt_q_sample}
\end{align}
Here, $h_0$ is the same as in~\eqref{eq:h0}. Next, we compute all the $h_t$'s and sample all the $z_t$'s by using $D_t + C_t$ instead of just $C_t$. The $h_t$'s for $t>0$ are computed as in~\eqref{eq:ht} using the generative network. All the $z_t$'s for $t > 0$ are drawn similar to~\eqref{eq:zt_sample} using $\draw_\phi$. Note that $\draw_\phi$ has access to the internal states of the generative $\draw_\theta$ network. This has been omitted in~\eqref{eq:zt_q_sample} for brevity but is described in the following sections.

\subsubsection{Basic Building Blocks}
\begin{asparaenum}
\item \textbf{Representation Network:} The representation network takes an image-viewpoint pair and summarizes the scene as a 3D tensor. Multiple such representations are combined in an order-invariant fashion by summing or averaging. We use the Tower Network as described in \cite{eslami2018neural}.
\begin{align*}
    D &= \{(\bx_1, \mathbf{y}_1), (\bx_2, \mathbf{y}_2), \ldots (\bx_m, \mathbf{y}_m)\}\\
    R_D &= \frac{1}{m}\sum_{i=1}^m \repnet{(\bx_i, \mathbf{y}_i)}
\end{align*}
Here, $D$ is a set of image-viewpoint pairs and $R_D$ is its computed representation.
\item \textbf{Convolutional LSTM Cell:} A standard LSTM Cell where all fully-connected layers are substituted for convolutional layers. 
\begin{align*}
    (h_{i+1}, c_{i+1}) \longleftarrow \convlstm{(\mathbf{x}_i, h_i, c_i)}
\end{align*}
where $h_i$ is the output of the cell and $c_i$ is the recurrent state of the ConvLSTM.
\end{asparaenum}

\subsubsection{Renderer $p(\by|\bz, \mathbf{h}, \mathbf{x})$}
The input to the renderer is the scene information contained in the latent $\bz$ and deterministic state $\mathbf{h}$ along with the camera viewpoint $\bx$. The output is the generated image $\by$. The renderer is deterministic and iterative where each iteration updates the image canvas as follows.

\begin{align*}
    \mathbf{e}^{(i)} &\leftarrow \encoder(\mathbf{y}^{(i)})\\
    (\mathbf{d}^{(i+1)}, \mathbf{c}^{(i+1)}) &\leftarrow \convlstm{(\mathbf{e}^{(i)}, \mathbf{d}^{(i)}, \mathbf{c}^{(i)}, \bx, \mathbf{h}, \mathbf{z})} \\
    \mathbf{y}^{(i+1)} &\leftarrow \mathbf{y}^{(i)} + \decoder{(\mathbf{d}^{(i+1)})}
\end{align*}
Here, $\bx^{(i)}$ is the canvas at the $i^{\supth}$ iteration and the $\mathbf{d}^{(i)}$ and $\mathbf{c}^{(i)}$ are the hidden state and the cell state of the convolutional LSTM, respectively. The number of iterations is a model parameter taken as 6.\\

Next, we describe the details of the encoder and decoder used above.
\begin{asparaenum}
\item \textbf{Encoder:} Details are shown in the Figure \ref{fig:renderer_enc}.
\begin{figure}[h!]
    \centering
    \includegraphics{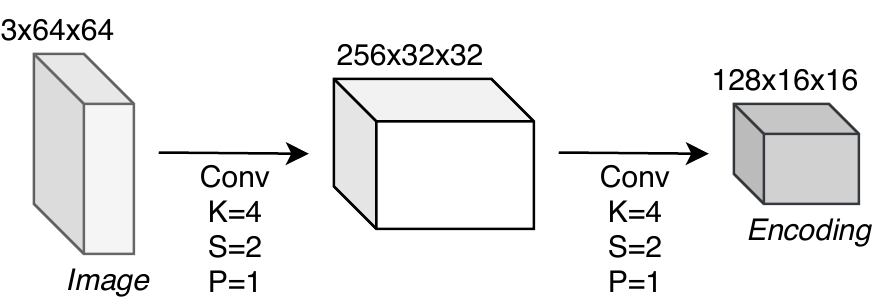}
    \caption{Encoder network has two convolutional layers. After each layer, ReLU non-linearity is applied.}
    \label{fig:renderer_enc}
\end{figure}

\item \textbf{Decoder:} Details are shown in the Figure \ref{fig:renderer_dec}.
\begin{figure}[h!]
    \centering
    \includegraphics{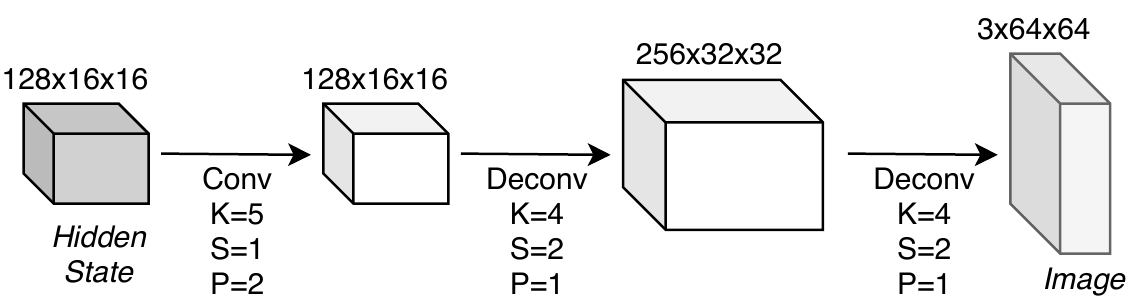}
    \caption{Encoder network has one convolutional layer and two transposed convolutional layers. After each layer except the last, ReLU non-linearity is applied.}
    \label{fig:renderer_dec}
\end{figure}
\end{asparaenum}

\subsubsection{Updating the deterministic state $\mathbf{h}_t$}
For any $t$, the deterministic state $\mathbf{h}_t$ summarizes all the previous latent states $\bz_{<t}$. This deterministic state is updated using a convolutional LSTM. The update may be described as follows.
\begin{align*}
    (\mathbf{h}_{t+1}, \mathbf{c}_{t+1}) &\leftarrow \convlstm{(\bz_t, \mathbf{a}_t, \mathbf{h}_{t}, \mathbf{c}_{t})}
\end{align*}
Here, $\mathbf{c}_{t}$ is the LSTM's internal cell state and $\mathbf{a}_t$ is the action received at time $t$.

\subsubsection{Sampling the latent $\bz_t$ using $p(\bz_t|\mathbf{h}_t, \mathbf{a}_t)$}

\begin{figure}[h!]
    \centering
    \begin{subfigure}{0.49\textwidth}
        \centering
        \includegraphics[width=0.6\linewidth]{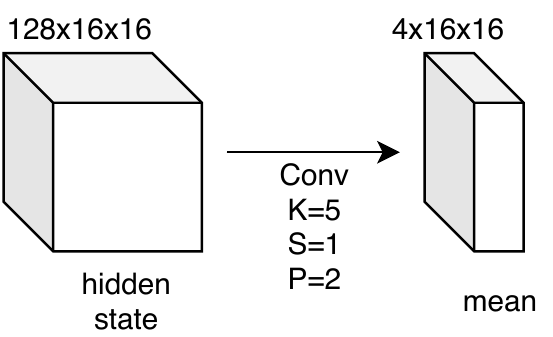}
    \end{subfigure}
    \begin{subfigure}{0.49\textwidth}
        \centering
        \includegraphics[width=0.6\linewidth]{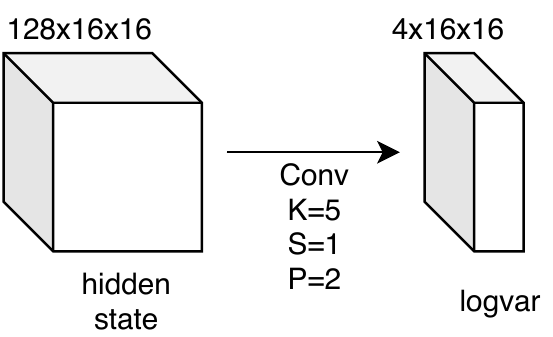}
    \end{subfigure}
    \caption{Computing sufficient statistics from the RNN hidden state of the auto-regressive density.}
    \label{fig:reparam}
\end{figure}

The sampling of latents, like CGQN \citep{kumar2018consistent}, is done using a DRAW-like auto-regressive density. Assume that \begin{enumerate*}[label=\itshape\alph*\upshape)]
\item $\mathbf{h}$ is the deterministic state,
\item $\mathbf{a}$ is the action provided,
\item $C$ is the context encoding provided at the current time-step and
\item $D$ is the target encoding provided at the current time-step.
\end{enumerate*}
\paragraph{Generation} This procedure is described in the following equations.
\begin{flalign}
(\hat{h}_0^p, \hat{c}_0^p) &\leftarrow \text{learned parameters} && \text{(Initial RNN state for generation)}\\
(\hat{h}_l^p, \hat{c}_l^p) &\leftarrow \rnn_\theta(z_t^{l-1}, \hat{h}_{l-1}^p, \hat{c}_{l-1}^p, \mathbf{h}, \mathbf{a}, C) && \text{(Update rule for generative RNN)} \label{eq:autoreg_p_h}\\
(\mu^l, \sigma^l) &\leftarrow \suffstats_\theta{(\hat{h}_l^p)} && \text{(See Fig.~\ref{fig:reparam})}\\
z^l &\sim \mathcal{N}(\mu^l, \sigma^l) && \text{(Sample the latent at current DRAW step)}
\end{flalign}
\paragraph{Inference} The inference procedure performs a similar sampling of the $z^l$'s but while having access to the hidden state of the generative RNN computed in~\eqref{eq:autoreg_p_h}. This procedure is described in the following equations.
\begin{flalign}
(\hat{h}_0^p, \hat{c}_0^p) &\leftarrow \text{learned parameters} && \text{(Initial RNN state for generation)}\\
(\hat{h}_0^q, \hat{c}_0^q) &\leftarrow \text{learned parameters} && \text{(Initial RNN state for inference)}\\
(\hat{h}_l^q, \hat{c}_l^q) &\leftarrow \rnn_\theta(z_t^{l-1},\hat{h}_{l-1}^q, \hat{h}_{l-1}^p, \hat{c}_{l-1}^q, \mathbf{h}, \mathbf{a}, D) && \text{(Update rule for inference RNN)} \label{eq:autoreg_q_h}\\
(\mu^l, \sigma^l) &\leftarrow \suffstats_\theta{(\hat{h}_l^q)} && \text{(See Fig.~\ref{fig:reparam})}\\
z^l &\sim \mathcal{N}(\mu^l, \sigma^l) && \text{(Sample the latent at current DRAW step)}\\
(\hat{h}_l^p, \hat{c}_l^p) &\leftarrow \rnn_\theta(z_t^{l-1}, \hat{h}_{l-1}^p, \hat{c}_{l-1}^p, \mathbf{h}, \mathbf{a}, C) && \text{(Update rule for the generative RNN)}
\end{flalign}
\newpage
\subsubsection{Hyper-Parameters}
In this sub-section, we describe the hyper-parameters used in our training.
\begin{table}[h!]
\centering
\begin{tabular}{@{}lll@{}}
\toprule
Parameter                     & 3D Tasks                             & 2D Tasks        \\ \midrule
Image Width/Height            & 64                                   & 64              \\
Image Channels                & 3                                    & 3               \\
Latent Width/Height           & 16                                   & 16              \\
Renderer Image Encoding Depth & 128                                  & 128             \\
ConvLSTM Hidden State Depth   & 128                                  & 128             \\
Context Representation Depth  & 256                                  & 256             \\
SSM Transition State Depth    & 108                                  & 108             \\
Training Batch-Size           & 4                                    & 4               \\
Latent Depth per DRAW step    & 4                                    & 4               \\
Action Input Embedding        & One-hot                              & N/A  \\
Number of DRAW steps          & 6                                    & 6               \\
Learning Rate                 & $10^{-5}$                              & $10^{-5}$          \\
Viewpoint Size                & 3                                    & 2               \\
RGB Distribution                    & Gaussian                      & Gaussian \\
RGB $\sigma^2$           & 2.0                               & 2.0           \\
Maximum context per time-step & 4                                    & 4 \\
\bottomrule
\end{tabular}
\end{table}
 \paragraph{Posterior Dropout} requires that we randomly choose between using $P_\ta$ or $Q_\phi$. The choice was made using a Bernoulli coin-toss with probability $0.3$ (for $Q_\phi$) at every time-step of each episode for each training iteration. Furthermore, the training of any task was first initiated without the posterior dropout ELBO i.e.~ with $\alpha =0$. The posterior dropout ELBO was turned on, i.e.~setting $\alpha=1$, after the reconstruction loss using the SNP ELBO had saturated. This is done to avoid conflict in the training of the encoder network due to two competing reconstruction losses from the two ELBOs in the initial stages of the training.
 \label{ax:hyperparam}

\subsection{GQN Baseline}
\label{ax:gqn_baseline}
Here, we provide some salient details of our implementation the GQN baseline.
\begin{enumerate*}[label=\itshape\alph*\upshape)]
\item In environments with actions, the query is a concatenation of the camera viewpoint and the RNN encoding of the action sequence up to that time-step. This RNN encoding has size 32. In action-less environments, $t$ as a normalized scalar concatenated to the camera viewpoint. 
\item We encode contexts (or targets) from multiple time-steps using sum-pooling as in original GQN.
\item During generation, TGQN cannot observe contexts from future time-steps. So for fair comparison at generation time, we also provide GQN with an encoding of contexts only up to the time-step that we are interested in querying.
\end{enumerate*}

\section{Dataset Additional Details}

\subsection{Gaussian Processes Dataset} \label{ax:1d_gp}

In each episode of task (a) and (b), the hyper-parameters of length-scale $l \in [0.7,1.2]$ and kernel-scale $\sigma \in [1.0, 1.6]$ are randomly drawn at $t=0$. In the task (c), $l$ and $\sigma$ are drawn from ranges $[1.2, 1.9]$ and $[1.6, 3.1]$, respectively. Similarly, the linear dynamics $\Delta l \in [-0.03,0.03]$ and $\Delta \sigma \in[-0.05,0.05]$ are also drawn randomly at $t=0$. To perform transitions, we execute $l+\Delta l$ and $\sigma + \Delta \sigma$ and add a small Gaussian noise at each time-step.

For task (a) and (b), the number of context and target are drawn randomly from $n \in [5,50]$ and $m \in [0, 50-n]$ whenever a non-empty context is being provided else $n=0$ and $m\in [0,50]$. For task (c), $n$ is 1 and $m$ is in $[0,10-n]$ whenever a non-empty context is being provided else $n=0$ and $m\in [0,10]$.

\subsection{2D Color Shapes Dataset}
\label{ax:2d_colorshapes}
The canvas and object sizes are $130\times130$ and $38\times38$, respectively. Speed of each object is 13 pixels per time-step and the initial direction is randomly chosen. The bouncing behaviour is modeled the same way as in the moving MNIST dataset \citep{srivastava2015unsupervised}. Shapes can be triangles, squares or circles and their colors can be red, magenta, blue, cyan, green or yellow. Here, we provide the fixed rule that we use to decide which object covers the other in case of an overlap.
\begin{itemize}
    \item Green or yellow cover red and magenta.
    \item Red or magenta cover blue and cyan.
    \item Magenta covers red.
    \item Cyan covers blue.
    \item Yellow covers green.
\end{itemize}

In this task, we pick the patch location (viewpoint) uniformly on the canvas. In the prediction regime, in each of the first 5 time-steps, we randomly decide the context set size $n$ uniformly in range $[1,5]$ and the target size $m$ is then taken as the number of remaining observations $20-n$. In the tracking regime, $n$ at each time-step is chosen in the range $[0,2]$ the remaining observations are used as the target.

\subsection{3D Environment Details}
\label{ax:mujoco}
We used the MuJoCo framework and the OpenAI Gym toolkit \citep{mordatch2015ensemble, brockman2016openai} to generate the 3D datasets. For training, we created 50,000 episodes where each episode contains 10 time-steps and each time-step contains 20 images. Therefore, the training is performed on 10 million images. For testing and evaluation, datasets containing 10,000 episodes with 30 time-steps each were separately generated. 

Actions at each time-step are uniformly randomly picked. If an action leads the object outside the arena, the action is re-picked until it doesn't. At each time $t$, we take 20 random camera angles in $[0,2\pi)$ and we use a part of it as context and leave the remaining as target. In the prediction regime, in each of the first 5 time-steps, we randomly decide the context set sizes uniformly in range $[1,5]$. In the tracking regime, $n$ at each time-step is chosen in the range $[0,2]$ the remaining observations are used as the target.

\end{appendices}

\end{document}